\documentclass[letterpaper]{article} 
\usepackage{aaai24}  
\usepackage{times}  
\usepackage{helvet}  
\usepackage{courier}  
\usepackage[hyphens]{url}  
\usepackage{graphicx} 
\usepackage{natbib}  
\usepackage{caption} 
\usepackage[linesnumbered,ruled,vlined]{algorithm2e}
\SetKwInOut{Parameter}{Parameter}

\usepackage{amsmath}
\usepackage{amssymb}
\usepackage{mathtools}
\usepackage{amsthm}
\newtheorem{assumption}{Assumption}

\newtheorem{theorem}{Theorem}

\newtheorem{corollary}{Corollary}
\newtheorem{example}{Example}

\newtheorem*{remark}{Remark}

\usepackage[capitalize]{cleveref}
\usepackage{bm}
\usepackage{dsfont}
\usepackage{multirow}
\usepackage{booktabs}
\usepackage{subfigure}

\def \cX {\mathcal{X}}
\def \cY {\mathcal{Y}}

\def \cR {\mathcal{R}}
\def \cL {\mathcal{L}}
\def \cS {\mathcal{S}}

\def \cT {\mathcal{T}}

\def \bbR {\mathbb{R}}
\def \bbE {\mathbb{E}}

\def \ni {m}
\def \nf {d}
\def \nc {c}
\def \ne {n}

\def \by {\bar{y}}
\def \bY {\bar{Y}}
\def \hp {\hat{p}}

\def \ABL {\mathrm{NeSy}}
\def \TL {\mathrm{TL}}
\def \L {\mathrm{L}}

\def \wy {\widetilde{y}}

\def \wQ {\widetilde{Q}}
\def \wq {\widetilde{q}}
\def \wc {\widetilde{c}}

\def \ind {\mathds{1}}

\def \ConjEq {\mathtt{ConjEq}}
\def \HED {\mathtt{HED}}
\def \Conjunction {\mathtt{Conjunction}}
\def \Addition {\mathtt{Addition}}
\def \DNF {\mathtt{DNF}}
\def \CNF {\mathtt{CNF}}

\def \ABLRAND {\textsc{Rand}}
\def \ABLMAXP {\textsc{MaxP}}
\def \ABLMIND {\textsc{MinD}}
\def \ABLAVG {\textsc{Avg}}
\def \ABLTL {\textsc{TL}}

\newcommand*{\img}[1]{%
    \raisebox{-.2\baselineskip}{%
        \includegraphics[
        height=0.85\baselineskip,
        width=0.85\baselineskip,
        keepaspectratio,
        ]{#1}%
    }%
  }

\title{Deciphering Raw Data in Neuro-Symbolic Learning with Provable Guarantees}
\author{
    Lue Tao,\textsuperscript{\rm 1,2}
    Yu-Xuan Huang,\textsuperscript{\rm 1,2}
    Wang-Zhou Dai,\textsuperscript{\rm 1,3}
    Yuan Jiang\textsuperscript{\rm 1,2}
}
\affiliations{
    \textsuperscript{\rm 1}National Key Laboratory for Novel Software Technology, Nanjing University, China\\
    \textsuperscript{\rm 2}School of Artificial Intelligence, Nanjing University, China\\
    \textsuperscript{\rm 3}School of Intelligence Science and Technology, Nanjing University, China\\
    \{taol, huangyx, daiwz, jiangy\}@lamda.nju.edu.cn
}

\begin{document}

\maketitle

\begin{abstract}
Neuro-symbolic hybrid systems are promising for integrating machine learning and symbolic reasoning, where perception models are facilitated with information inferred from a symbolic knowledge base through logical reasoning. Despite empirical evidence showing the ability of hybrid systems to learn accurate perception models, the theoretical understanding of learnability is still lacking. Hence, it remains unclear why a hybrid system succeeds for a specific task and when it may fail given a different knowledge base. In this paper, we introduce a novel way of characterising supervision signals from a knowledge base, and establish a criterion for determining the knowledge's efficacy in facilitating successful learning. This, for the first time, allows us to address the two questions above by inspecting the knowledge base under investigation. Our analysis suggests that many knowledge bases satisfy the criterion, thus enabling effective learning, while some fail to satisfy it, indicating potential failures. Comprehensive experiments confirm the utility of our criterion on benchmark tasks.
\end{abstract}

\section{Introduction}

Integrating machine learning and symbolic reasoning is a holy grail challenge in artificial intelligence.
This pursuit has attracted much attention over the past decades~\cite{garcez2002neural, getoor2007introduction, russell2015unifying, de2021statistical, hitzler2022neuro}, leading to fruitful developments such as probabilistic logic programing \cite{de2015probabilistic} and statistical relational artificial intelligence \cite{raedt2016statistical}.

In recent years, great progress has been made in neuro-symbolic methods, equipping symbolic systems with the ability to perceive sub-symbolic data. 
One intriguing finding in these hybrid systems is that the perception performance of initialised classifiers can be significantly enhanced through \textit{abduction}, a.k.a.~\textit{abductive reasoning} \cite{dai2019bridging, li2020closed}.
Moreover, it has been shown that accurate classifiers can be learned from scratch without relying on fully labelled data, given appropriate objectives and knowledge bases \cite{xu2018semantic, manhaeve2018deepproblog, tsamoura2021neural, dai2021abductive}.

These advances highlight the value of symbolic reasoning in many learning tasks.
However, not all symbolic knowledge helps improve learning performance; there are failures in practice~\cite{cai2021abductive, marconato2023neuro, marconato2023not}.
More importantly, the theoretical underpinnings that drive these empirical successes or failures remain elusive, which may hinder the adoption of neuro-symbolic methods in other applications.
In particular, it is unclear why such a hybrid learning system works for a specific task and when it may fail given a different knowledge base.

\begin{figure}[t]
\centering
\includegraphics[width=0.9\columnwidth]{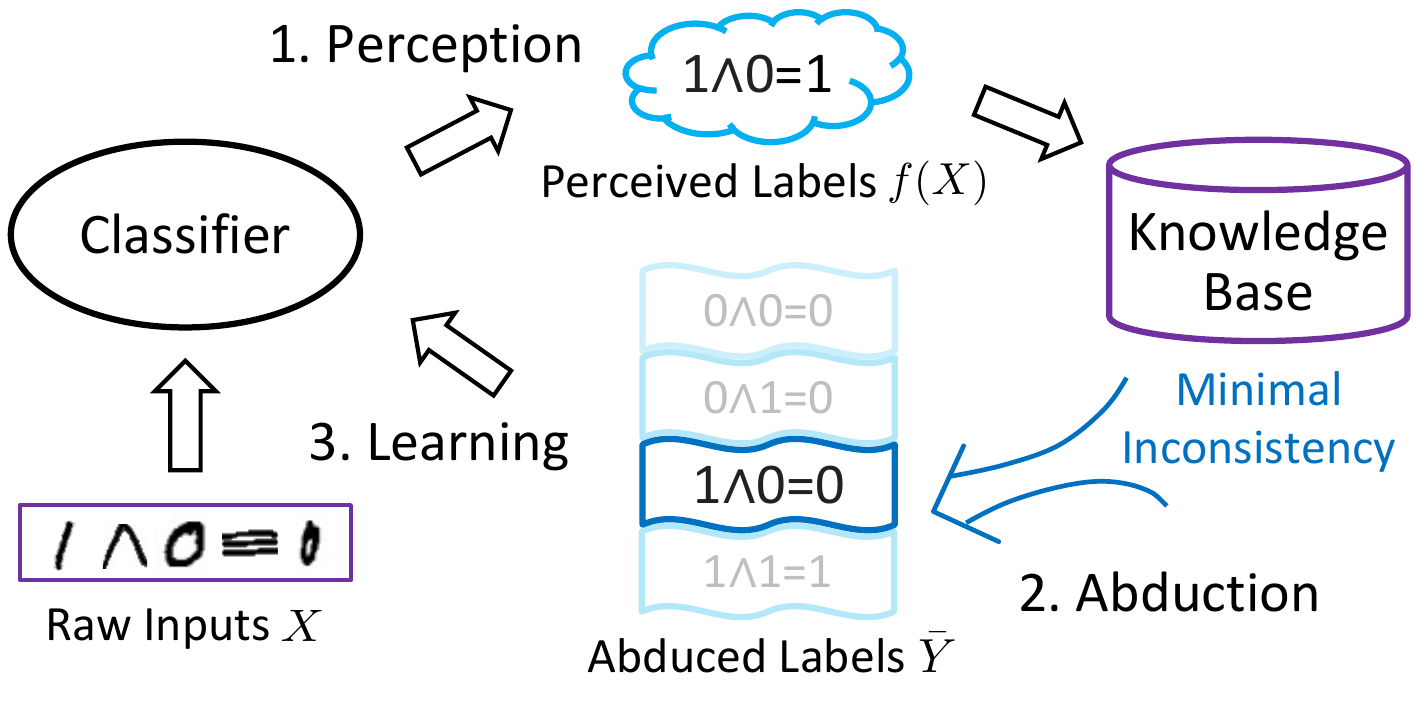}
\vspace{-3px}
\caption{
An illustration of the hybrid learning framework.
First, raw data such as handwritten equations are perceived by a classifier.
Next, the perceived labels are revised via logical abduction under the principle of minimal inconsistency.
Finally, the abduced labels are used to update the classifier.
}
\label{fig:abl-paradigm}
\vspace{-8px}
\end{figure}

In this paper, we address the above questions under the framework of \textit{abductive learning} (ABL), an expressive and representative paradigm of hybrid systems \cite{DBLP:journals/chinaf/Zhou19, zhou2022abductive}.
As illustrated in \cref{fig:abl-paradigm}, a hybrid learning system usually involves the perception of raw inputs using a classifier, followed by an abductive reasoning module~\cite{kakas1992abductive} that aims to correct the wrongly perceived labels by minimising the inconsistency with a given symbolic knowledge base.

\paragraph{Contributions.}
We present a theoretical analysis illuminating the factors underlying the success of hybrid systems.
Our analysis is based on a key insight: the objective of popular hybrid methods implies a secret objective that finely characterises supervision signals from a given knowledge base.
Formally, we show that the objective of minimising the inconsistency with a knowledge base, under reasonable conditions, is equivalent to optimising an upper bound of another objective denoted by L-Risk, which contains a probability matrix relating the ground-truth labels and their locations in the ground atoms of the knowledge base. This indicates that hybrid methods can make progress by mitigating the L-Risk. 
Further, we show that if the matrix in the objective is full-rank, the ground-truth values of the labels perceived from raw inputs are guaranteed to be recovered.
In this manner, we establish a rank criterion that indicates the knowledge's efficacy in facilitating successful learning.

To our knowledge, this is the first study that attempts to provide a reliable diagnosis of the knowledge base in abductive learning prior to actual training. 
Our theoretical analysis offers practical insights: if the knowledge base meets the criterion, it can facilitate learning; otherwise, it might fail and require further refinement to ensure successful learning. Comprehensive experiments on benchmark tasks with different knowledge bases validate the utility of the criterion. 
We believe that our findings are instrumental in guiding the integration of machine learning and symbolic reasoning.

\section{Preliminaries}

\subsubsection{Conventional Supervised Learning.}
Let $\cX \subseteq \bbR^\nf$ be the input space and $\cY = [\nc] = \{0, \ldots, c-1\}$ be the label space, where $\nc$ is the number of classes. 
In ordinary multi-class learning, each instance-label pair $\langle x, y \rangle \in \cX \times \cY$ is sampled from an underlying distribution with probability density $p(x, y)$, and the objective is to learn a mapping $h: \cX \rightarrow \bbR^\nc$ that minimises the expected risk over the distribution:
\begin{equation}
\label{eq:sl-risk}
\cR(h) = \bbE_{p(x, y)}  \ell(h(x), y) ,
\end{equation}
where $\ell: \bbR^\nc \times \cY \rightarrow \bbR$ is a loss function that measures how well the classifier perceives an input.
The predicted label of the classifier is represented by $f(x) = \arg\max_{i\in\cY}h_i(x)$, where $h_{i}(x)$ is the $i$-th element of $h(x)$.

\subsubsection{Neuro-Symbolic Learning.} 
In neuro-symbolic (NeSy) learning systems, it is common to assume that raw inputs $X = [x_0, x_1, \ldots, x_{\ni-1}]$ are given, while their ground-truth labels $Y = [y_0, y_1, \ldots, y_{\ni-1}]$ are not observable. Instead, we only know that the logical facts grounded by the labels are compatible with a given knowledge base. Specifically, we have $B \cup Y \models \tau$, where $B$ is a knowledge base consisting of first-order logic rules, $\models$ denotes logical entailment, and $\tau$ is a target concept in a concept space $\cT$. 

\begin{figure}[t]
\vspace{-3px}
\centering
\includegraphics[width=0.95\columnwidth]{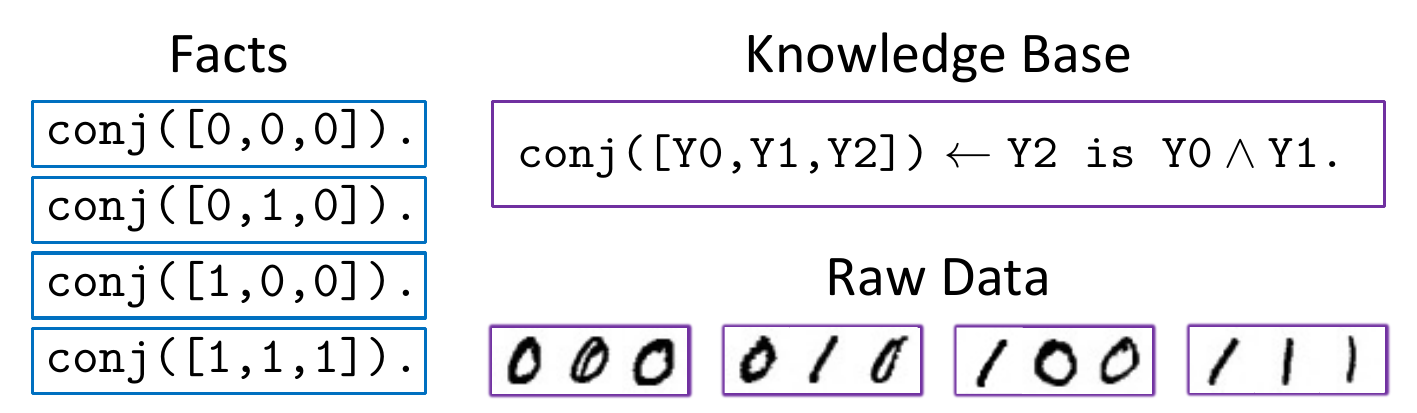}
\vspace{-3px}
\caption{
Illustration of the knowledge base about conjunction, the facts abduced from the knowledge base, and the raw inputs corresponding to the target concept ``$\mathtt{conj}$''.
}
\label{fig:motivating-example}
\vspace{-8px}
\end{figure}

\begin{example}
\label{exp:conj}
\normalfont
Consider a binary classification task with label space $\cY=\{0,1\}$ and concept space $\cT=\{\mathtt{conj}\}$, which contains a single target concept ``$\mathtt{conj}$''. 
Each sequence $X$ consists of three raw inputs, whose ground-truth labels $Y$ are unknown but satisfy the logical equation $y_0 \wedge y_1 = y_2$.
\cref{fig:motivating-example} illustrates the logical facts abduced from the knowledge base and several sequences of raw inputs in this task.
\end{example}

In \cref{exp:conj}, when observing a sequence of raw inputs $X = [\img{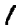}, \img{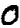}, \img{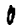}]$ with the target concept ``$\mathtt{conj}$'', the classifier should learn to perceive the inputs so that the logical equation $f(\img{fig/inverted_eq010-2.png}) \wedge f(\img{fig/inverted_eq010-1.png}) = f(\img{fig/mnist_0_2094.png})$ holds.
In general, denote by $p(X, \tau)$ the underlying distribution of the input sequence $X \in \cX^\ni$ and the target concept $\tau \in \cT$, the objective is to learn a mapping $h: \cX \rightarrow \bbR^\nc$ that minimises the inconsistency between the classifier and the knowledge base:
\begin{equation}
\label{eq:mi-risk}
\cR_{\ABL}(h) = \bbE_{p(X, \tau)}  \cL(X, \bY; h) , \text{ s.t. } B\cup \bY \models \tau,
\end{equation}
where $\cL(X, \bY; h) = \frac{1}{\ni} \sum_{k=0}^{\ni-1}  \ell(h(x_k), \by_k) $, and $\bY=[\by_0, \by_1, \ldots, \by_{\ni-1}]$ denotes the \textit{abduced labels} that are consistent with the knowledge base $B$.
The abduced labels $\bY$ are inferred through \textit{abduction}, a basic form of logical reasoning that seeks the most likely explanation for observations based on background knowledge~\cite{sanders1955abduction, simon1971human, garcez2007abductive}.
Often, there are multiple candidates for abduced labels~\cite{dai2019bridging}, e.g., both $0 \wedge 1 = 0$ and $1 \wedge 0 = 0$ are correct equations. 
Hence, various heuristics have been proposed to guide the search for the most likely labels from the candidate set $\cS(\tau) = \{ \bY \in \cY^\ni \mid B \cup \bY \models \tau \}$.
For example, \citet{cai2021abductive} constrained the Hamming distance between the abduced labels $\bY$ and the predicted labels $f(X)$,\footnote{With a slight abuse of notation, the predicted labels of raw inputs are denoted by $f(X) = [f(x_0), f(x_1), \ldots, f(x_{\ni-1})]$.} and \citet{dai2021abductive} chose to pick the most probable labels based on the likelihood $p(\bY|X) = \prod_{k=0}^{\ni-1} p(\by_k|x_k)$, where $p(\by|x)$ is approximated by the output of softmax function, i.e., $\hp(\by|x) = \exp({h_{\by}(x)}) / \sum_{i=0}^{\nc-1} \exp(h_i(x))$.
The overall pipeline of these algorithms is described in \cref{app:algorithms}.


\section{Theoretical Analysis}
\label{sec:theoretical-analysis}

Previous studies have showcased the practicality of neuro-symbolic learning systems---the objective of minimal inconsistency empirically yields classifiers adept at accurately predicting labels.
In this section, we aim to disclose the ingredients of success from a theoretical perspective. 
We begin by considering a simple yet representative task.
This motivates us to formulate a novel way of characterising supervision signals from a given knowledge base, and provide conditions under which the signals are sufficient for learning to succeed.
Specifically, 
we show that the objective in \cref{eq:mi-risk} essentially addresses an upper bound of a location-based risk, whose minimisers are guaranteed to recover ground-truth labels when a rank criterion is satisfied.

\subsection{Location Signals}
\label{sec:motivating-example}

Let us first consider the sequence $[\img{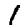},\img{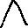},\img{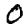},\img{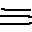},\img{fig/100-5.png}]$ in \cref{fig:abl-paradigm}. It forms a correct equation only if the penultimate instance $\img{fig/100-4.png}$ is labelled as ``equal-sign''.  
This illustrates that the label of an instance could be determined by its location.
However, challenges remain, e.g., the label of the last instance $\img{fig/100-5.png}$ is not determined by its location. To resolve this, our intuition is that the last instance is more likely to be $0$ than $1$, since there are four possible cases for a correct equation: $[0,\wedge, 0, =, 0]$, $[0,\wedge, 1, =, 0]$, $[1,\wedge, 0, =, 0]$, and $[1,\wedge, 1, =, 1]$. In $3/4$ cases, the last instance is $0$.
This intuition is utilised as follows.

\begin{figure}[t]
\vspace{-3px}
\centering
\includegraphics[width=0.95\columnwidth]{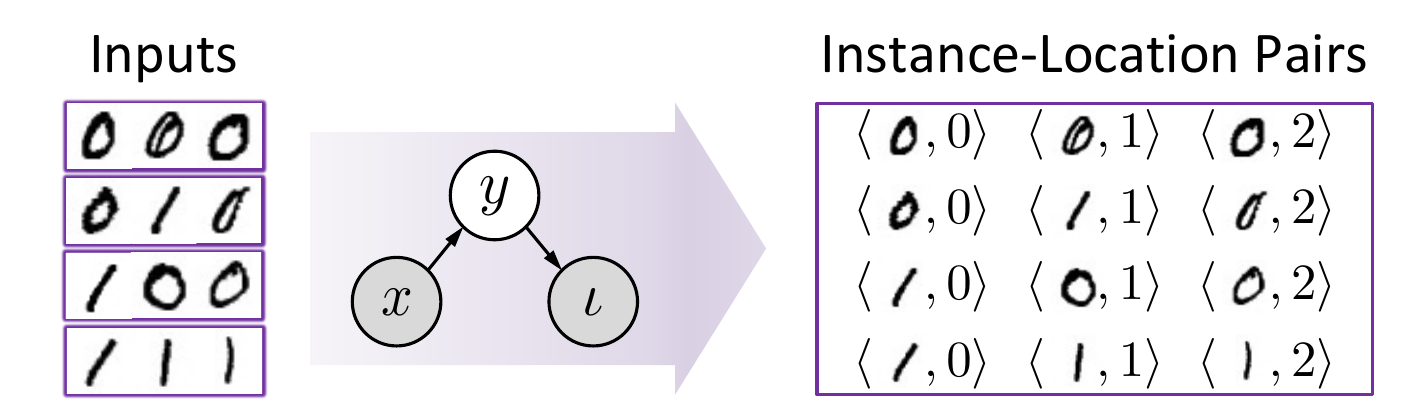}
\vspace{-3px}
\caption{
Illustration of the location signals in \cref{exp:conj}.
From the input sequences, we can observe instance-location pairs $\langle x, \iota \rangle$, while ground-truth labels $y$ are unobservable.
Intuitively, the instances of $y=0$ are more likely to occur at the $2$-th position, if candidate label probabilities are equal.
}
\label{fig:data-collection-example}
\vspace{-8px}
\end{figure}

Now, consider the task in \cref{exp:conj}. For each input sequence, there are four possibilities of abduced labels including $[0,0,0]$, $[0,1,0]$, $[1,0,0]$, and $[1,1,1]$.
This sequence-level information is crude, while our interest is still in instance-level prediction.
To this end, we would like to extract instance-level supervision signals from the sequences.
Specifically, in \cref{exp:conj}, each input sequence $[x_0, x_1, x_2]$ naturally yields three instance-location pairs $\{\langle x_{\iota}, \iota \rangle\}_{\iota=0}^{2}$, where $\iota$ denotes the location of an instance in the sequence.
\cref{fig:data-collection-example} illustrate the instance-location pairs.
In general, given $n$ sequences of unlabelled data $\{X^{(i)}\}_{i=0}^{\ne-1}$, each sequence $X^{(i)} = [x^{(i)}_0, x^{(i)}_1, \ldots, x^{(i)}_{\ni-1}]$ naturally yields $m$ instance-location pairs $\{\langle x^{(i)}_{\iota}, \iota \rangle \}_{\iota=0}^{\ni-1}$, and finally a total of $\ni\ne$ instance-location pairs $\{\langle x^{(i)}, \iota^{(i)} \rangle\}_{i=0}^{\ni\ne-1}$ are obtained.

Apparently, such instance-level signals are insufficient for predicting true labels $y$. With a sample of the instance-location pairs, we could only learn a mapping $q: \cX \rightarrow \bbR^\ni$ that estimates the underlying conditional probability $p(\iota | x)$. To address this issue, we choose to express $p(\iota | x)$ in terms of the desired conditional probability $p(y | x)$; that is, $\forall k \in [\ni]$,
\begin{equation}
\label{eq:simple-reformulation}
\textstyle
    p(\iota=k \mid x) = \sum_{j=0}^{\nc-1} Q_{jk} \cdot p(y=j \mid x),
\end{equation}
where $Q_{jk} = p(\iota=k | y=j)$ denotes the probability of the class $j$ occurring at the $k$-th position in a sequence.
Here, we assume $p(\iota|y,x)=p(\iota|y)$. This means that $\iota$ is independent of $x$ given $y$, analogous to the “missing completely at random” assumption that is often made in learning with missing values \cite{little1987statistical, elkan2008learning}.

In practice, we let $q(x) = Q^\top g(x)$ and interpret $g(x)$ as probabilities via $g_{j}(x) = \exp({h_{j}(x)}) / \sum_{i=0}^{\nc-1} \exp(h_i(x))$, $\forall j \in [\nc]$.
Consequently, if $q(x)$ learns to predict the probability $p(\iota | x)$, then $g(x)$ serves to estimate the probability $p(y | x)$. This can be achieved by minimising the following location-based risk~(L-Risk) with appropriate loss functions:
\begin{equation}
\textstyle
\label{eq:l-risk} 
\cR_{\L}(h) = \bbE_{p(x, \iota)}  \ell(q(x), \iota).
\end{equation}

While the above indicates that it is possible to decipher ground-truth labels, the reliance on the knowledge of label distribution is indispensable. Indeed, we find a frequently-used data generation process in previous practices~\cite{cai2021abductive, dai2021abductive, huang2021fast}, where the labels of input sequences are implicitly assumed to be uniform over the candidate set $\cS(\tau)$. 
This means that the prior probabilities of candidate labels are equal. Under this assumption, the probability $Q_{jk}$ in \cref{eq:simple-reformulation} can be computed
as $\sum_{Y\in\cS(\tau)} \ind(y_k=j) / \sum_{Y\in\cS(\tau)} \sum_{k=0}^{\ni-1} \ind(y_k=j)$, where $\ind(\cdot)$ is the indicator function.
For the task in \cref{exp:conj}, the uniform assumption implies that the probabilities of candidate labels are $p([0,0,0]) = p([0,1,0]) = p([1,0,0]) = p([1,1,1]) = 1/4$, and that the elements in the probability matrix $Q$ are $Q_{00} = 2/7, Q_{01} = 2/7, Q_{02} = 3/7$ and $Q_{10} = 2/5, Q_{11} = 2/5, Q_{12} = 1/5$.


In what follows, we will establish connections between the location-based risk and the objective of popular hybrid methods by utilising and relaxing the uniform assumption.

\subsection{Upper Bound}
\label{eq:upper-bound}

We first show that under reasonable conditions, the objective of minimal inconsistency in \cref{eq:mi-risk} is an upper bound of the L-Risk in \cref{eq:l-risk} up to an additive constant.

\begin{assumption}[Uniform Assumption]
\label{asp:uniform}
    $\forall Y \in \cS(\tau)$, $p(Y) = 1/|\cS(\tau)|$, where $\cS(\tau) = \{ Y \in \cY^\ni \mid B \cup Y \models \tau \}$.
\end{assumption}

\begin{theorem}
\label{thrm:simple-upper-bound}
Suppose that the uniform assumption holds and the concept space contains only one target concept $\tau$.
Let $a = \max_{i\in\cY} \{ \sum_{Y\in\cS(\tau)} \sum_{k=0}^{\ni-1} \ind(y_{k}=i) / |\cS(\tau)| \}$.
For any classifier $h$ and any knowledge base $B$, if the abduced labels $\bY$ are randomly selected from $\cS(\tau)$ and $\ell$ is the cross-entropy loss, then we have
$$
\textstyle
\cR_{\L}(h) \le \cR_{\ABL}(h) + C,
$$
where $C = \log a \le \log \ni$ is a constant.
\end{theorem}

The proof is given in \cref{app:proof-simple-upper-bound}. 
\cref{thrm:simple-upper-bound} illustrates that minimising the inconsistency with a given knowledge base in hybrid systems is equivalent to minimising an upper bound of the location-based risk. 
In addition, the bound is tight: equality can be achieved, for example, when $\cS(\tau)=\{[0,1], [1,0]\}$ and a uniformly random classifier $h$ such that for any $x$ and $y$, $\ell(h(x), y) = -\log \hp(y|x) = \log 2$.

\begin{remark}
\normalfont
Firstly, it is common in practice to collect a set of input sequences belonging to the same target concept. For example, the word recognition task in \citet{cai2021abductive} contains only one target concept ``$\mathtt{valid\_word}$'', and the equation decipherment experiment in \citet{huang2021fast} uses only correct equations.
Secondly, although many heuristics have been proposed to select the most likely labels from the candidate set, they all behave like random guessing in the early stages of training when the classifier is randomly initialised \cite{dai2021abductive}. 
Finally, the cross-entropy loss is commonly adopted in previous work \cite{dai2019bridging, cai2021abductive, huang2021fast}.
\end{remark}

Above, we have mainly focused on learning with a single target concept, which is a representative case due to its broad applications, and due to the insight it offers into the location-based signals.
Next, we will generalise our discussion to utilise multiple target concepts.
Evidently, these target concepts provide another source of supervision signals.

\begin{figure}[t]
\vspace{-3px}
\centering
\includegraphics[width=0.95\columnwidth]{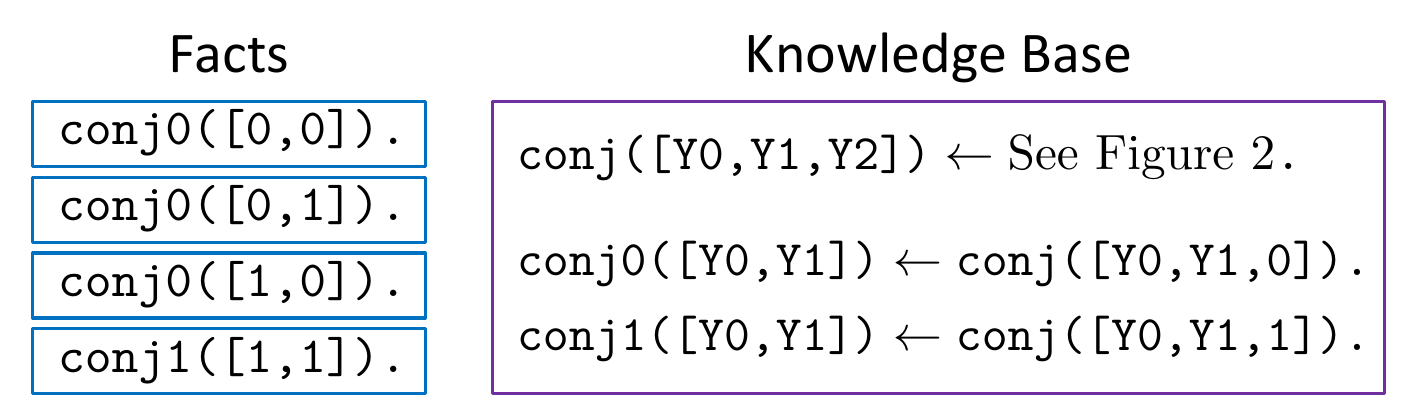}
\vspace{-3px}
\caption{
Illustration of the knowledge base about the target concepts ``$\mathtt{conj0}$'' and ``$\mathtt{conj1}$'', along with the corresponding facts abducted from the knowledge base.
}
\label{fig:multi-concepts-example}
\vspace{-8px}
\end{figure}

\begin{example}
\label{exp:conj01}
\normalfont
Consider a binary classification task with label space $\cY=\{0,1\}$ and concept space $\cT=\{\mathtt{conj0}, \mathtt{conj1}\}$.
Each sequence $X$ consists of two raw inputs, whose ground-truth labels $Y$ are unknown but satisfy the logical equations $y_0 \wedge y_1 = 0$ or $y_0 \wedge y_1 = 1$.
\cref{fig:multi-concepts-example} illustrates the logical facts abduced from the knowledge base in this task.
\end{example}

For each input sequence in \cref{exp:conj01}, three possibilities of candidate labels can be abduced from the target concept ``$\mathtt{conj0}$'', including $[0, 0]$, $[0, 1]$ and $[1,0]$, while the candidate labels become $[1,1]$ if the observed concept is ``$\mathtt{conj1}$''.
In order to simultaneously exploit this target-based information and the aforementioned location-based signals, we propose to construct a collection of instance-location-target triplets. 
Specifically, given $n$ input sequences paired with target concepts $\{\langle X^{(i)}, \tau^{(i)} \rangle\}_{i=0}^{\ne-1}$, each sequence naturally yields $\ni$ triplets $\{ 
\langle x^{(i)}_\iota, \iota, \tau^{(i)} \rangle \}_{\iota=0}^{\ni-1}$, and finally a set of $\ni \ne$ triplets $\{\langle x^{(i)}, \iota^{(i)}, \tau^{(i)} \rangle\}_{i=0}^{\ni\ne-1}$ are obtained.

In combination, the signals $\langle \iota^{(i)}, \tau^{(i)} \rangle$ in the triplets have $\wc = \ni \cdot |\cT|$ distinct values.
For conciseness, we use synthetic labels $\wy \in [\wc]$ to denote the signals $\langle \iota, \tau \rangle \in [\ni] \times \cT$. Then, the triplet set is represented as a set of instance-label pairs $\{\langle x^{(i)}, \wy^{(i)} \rangle\}_{i=0}^{\ni \ne - 1}$, from which we could learn a mapping $\wq: \cX \rightarrow \bbR^{\wc}$ that estimates the underlying conditional probability $p(\wy|x)$, i.e., $p(\iota, \tau|x)$.
Similar to before, we express $p(\wy|x)$ in terms of $p(y|x)$ for predicting ground-truth labels; that is, $\forall o \in [\wc]$,
\begin{equation}
\label{eq:general-reformulation}
\textstyle
    p(\wy=o \mid x) = \sum_{j=0}^{\nc-1} \wQ_{jo} \cdot p(y=j \mid x),
\end{equation}
where $\wQ_{jo} = p(\wy=o | y=j)$.
Without loss of generality, let $t\in\{0, \ldots, |\cT|-1\}$ and $o=tm+k$. Then, $\wQ_{jo}$ represents $p(\iota=k, \tau=t | y=j)$, which is the probability of the class $j$ occurring at the $k$-th position in a sequence of concept~$t$.
In practice, we let $\wq(x) = \wQ^{\top} g(x)$. If $\wq(x)$ learns to predict the probability $p(\wy|x)$, then $g(x)$ serves to estimate the probability $p(y|x)$. This can be achieved by minimising the following target-location-based risk (TL-Risk):
\begin{equation}
\textstyle
\label{eq:tl-risk} 
\cR_{\TL}(h) = \bbE_{p(x, \wy)}  \ell(\wq(x), \wy).
\end{equation}

Now we are ready to state the upper bound in the case of multiple target concepts.

\begin{theorem}
\label{thrm:upper-bound-general-case}
Given a data distribution $p(X, \tau)$ with any label distribution $p(Y)$. Let $b = \min_{\tau\in\cT} p(\tau)$.
For any classifier $h$ and any knowledge base $B$, if the abduced labels $\bY$ are randomly selected from $\cS(\tau)$ and $\ell$ is the cross-entropy loss, then we have
$$
\cR_{\TL}(h) \le \cR_{\ABL}(h) + C,
$$
where $C = \log\ni - \log b$ is a constant.
\end{theorem}

The proof is given in \cref{app:proof-upper-bound}. \cref{thrm:upper-bound-general-case} illustrates that minimising the inconsistency in hybrid systems is equivalent to minimising an upper bound of the target-location-based risk. 
Also, \cref{thrm:upper-bound-general-case} implies that the upper bound holds independent of the uniform assumption.
However, to render \cref{eq:tl-risk} practical, a premise for obtaining the value of the probability matrix $\wQ$ is still required.

To this end, we resort to a realistic data generation process described as follows.
Initially, instances in a sequence are generated independently from $p(x, y)$. This sequence is subsequently submitted to a labelling oracle, which generates a target concept by identifying a ground atom in its knowledge base that corresponds to the instances' labels. This procedure is repeated, eventually yielding a collection of sequences paired with their target concepts.
With this data generation process, it is straightforward to derive the sequence-level label density $p(Y)$ from the instance-level label density $p(y)$. 
More specifically, the label density for a sequence is given as the product of the individual instance-level densities., i.e., $p(Y=[\by_1, \ldots, \by_m]) = \prod_{k=0}^{\ni-1} p(y=\by_k)$.
Then, by following the derivation of generality in \citet{muggleton2023hypothesizing}, we obtain $p(Y|\tau) = p(\tau|Y)p(Y)/p(\tau)$,
where $p(\tau=t | Y=\bY)$ equals one if $B \cup \bY \models t$, otherwise it equals zero. 
Finally, the probability $\wQ_{jo}$ can be computed as $p(\tau,\iota|y) = p(y|\tau,\iota)p(\tau)p(\iota)/p(y)$, where $p(y=j | \tau=t, \iota=k) = \sum_{\bY \in \cS(t)} \ind(\by_{k} = j) p(Y=\bY | \tau=t)$.

\begin{remark}
\normalfont
The data generation process described above has been adopted in previous work, such as the ``addition'' task in \citet{manhaeve2018deepproblog} and the ``member'' task in \citet{tsamoura2021neural}. 
We also note that when the prior distribution on $y$ is uniform, the label density on $Y$ given a target concept $\tau$ derived from the above process is also uniform over the candidate set.
Thus, this data generation process favourably relaxes the uniform assumption.
\end{remark}

\subsection{Rank Criterion}

Equipped with the probability matrix, now we are ready to present a rank criterion that indicates the knowledge's efficacy in recovering the ground-truth labels of raw inputs.

\begin{theorem}
\label{thrm:rank-criterion}
If the probability matrix $\wQ$ has full row rank and the cross-entropy loss is used, then the minimiser $h^*_{\TL} = \arg\min_{h} \cR_{\TL}(h)$ recovers the true minimiser $h^* = \arg\min_{h} \cR(h)$, i.e., $h^*_{\TL} = h^*$.
\end{theorem}

The proof is given in \cref{app:proof-rank-criterion}. \cref{thrm:rank-criterion} means that the true values of the labels of raw inputs can be reliably recovered if the probability matrix has full row rank.


On one hand, \cref{thrm:simple-upper-bound,thrm:upper-bound-general-case} demonstrate that the objective of minimal inconsistency contributes to mitigating location-based risks, in which the supervision signals implied by a knowledge base are explicitly characterised into a probability matrix. 
On the other hand, \cref{thrm:rank-criterion} reveals that the minimiser of the TL-Risk can recover the true labels, provided that the probability matrix has full row rank.
In~short, these findings suggest that the objective of minimal inconsistency can succeed by mitigating the risk of predicting locations, with the rank criterion serving as an indicator of the knowledge's efficacy in facilitating learning.

We conclude this part by illustrating the use of the rank criterion for learning with one or more target concepts.

\begin{corollary}
\label{coro:rank-criterion}
If the probability matrix $Q$ has full row rank and the cross-entropy loss is used, then the minimiser $h^*_{\L} = \arg\min_{h} \cR_{\L}(h)$ recovers the true minimiser $h^* = \arg\min_{h} \cR(h)$, i.e., $h^*_{\L} = h^*$.
\end{corollary}

\begin{table*}[!t]
\centering
\begin{small}
\resizebox{1.0\textwidth}{!}{
\setlength{\tabcolsep}{3.6mm}{
\begin{sc}
\begin{tabular}{c|c|c|c|c|c|c}
\toprule
Task                            & Method     & MNIST            & EMNIST           & USPS             & Kuzushiji        & Fashion          \\ \midrule
\multirow{5}{*}{$\ConjEq$}      & $\ABLRAND$ & $99.91 \pm 0.06$ & $99.65 \pm 0.04$ & $99.33 \pm 0.16$ & $97.82 \pm 0.35$ & $98.40 \pm 0.11$ \\
                                & $\ABLMAXP$ & $99.94 \pm 0.04$ & $99.82 \pm 0.03$ & $99.20 \pm 0.00$ & $98.80 \pm 0.16$ & $99.39 \pm 0.12$ \\
                                & $\ABLMIND$ & $99.91 \pm 0.08$ & $99.84 \pm 0.07$ & $99.14 \pm 0.17$ & $98.91 \pm 0.17$ & $98.84 \pm 0.19$ \\
                                & $\ABLAVG$  & $99.85 \pm 0.10$ & $99.80 \pm 0.07$ & $99.30 \pm 0.17$ & $98.34 \pm 0.16$ & $98.62 \pm 0.21$ \\
                                & $\ABLTL$   & $99.92 \pm 0.05$ & $99.82 \pm 0.06$ & $99.25 \pm 0.08$ & $98.53 \pm 0.26$ & $98.77 \pm 0.06$ \\ \midrule
\multirow{5}{*}{$\Conjunction$} & $\ABLRAND$ & $99.91 \pm 0.06$ & $99.86 \pm 0.04$ & $99.30 \pm 0.13$ & $98.79 \pm 0.13$ & $99.00 \pm 0.27$ \\
                                & $\ABLMAXP$ & $99.93 \pm 0.04$ & $99.81 \pm 0.02$ & $99.20 \pm 0.00$ & $98.62 \pm 0.15$ & $99.05 \pm 0.09$ \\
                                & $\ABLMIND$ & $99.94 \pm 0.02$ & $99.79 \pm 0.02$ & $99.20 \pm 0.00$ & $98.74 \pm 0.10$ & $99.08 \pm 0.10$ \\
                                & $\ABLAVG$  & $99.94 \pm 0.02$ & $99.85 \pm 0.03$ & $99.30 \pm 0.13$ & $98.68 \pm 0.33$ & $99.23 \pm 0.13$ \\
                                & $\ABLTL$   & $99.94 \pm 0.02$ & $99.83 \pm 0.04$ & $99.20 \pm 0.00$ & $98.87 \pm 0.18$ & $99.30 \pm 0.04$ \\ \midrule
\multirow{5}{*}{$\Addition$}    & $\ABLRAND$ & $92.01 \pm 0.93$ & $92.94 \pm 1.45$ & $90.96 \pm 1.04$ & $73.18 \pm 0.71$ & $79.08 \pm 2.61$ \\
                                & $\ABLMAXP$ & $96.40 \pm 4.04$ & $95.09 \pm 5.20$ & $94.29 \pm 0.27$ & $90.00 \pm 0.27$ & $87.34 \pm 2.93$ \\
                                & $\ABLMIND$ & $98.32 \pm 0.04$ & $98.61 \pm 0.06$ & $94.61 \pm 0.17$ & $90.85 \pm 0.26$ & $88.40 \pm 0.62$ \\
                                & $\ABLAVG$  & $94.90 \pm 0.39$ & $95.71 \pm 0.42$ & $93.22 \pm 0.30$ & $80.94 \pm 0.62$ & $84.43 \pm 0.92$ \\
                                & $\ABLTL$   & $98.00 \pm 0.14$ & $98.41 \pm 0.05$ & $94.68 \pm 0.20$ & $90.04 \pm 0.32$ & $88.38 \pm 0.25$ \\ \midrule
\multirow{5}{*}{$\HED$}         & $\ABLRAND$ & $99.89 \pm 0.02$ & $99.71 \pm 0.12$ & $99.25 \pm 0.23$ & $97.68 \pm 0.70$ & $98.43 \pm 0.55$ \\
                                & $\ABLMAXP$ & $99.90 \pm 0.02$ & $99.77 \pm 0.02$ & $99.23 \pm 0.05$ & $98.55 \pm 0.08$ & $99.33 \pm 0.10$ \\
                                & $\ABLMIND$ & $99.87 \pm 0.07$ & $99.77 \pm 0.02$ & $99.21 \pm 0.00$ & $98.61 \pm 0.21$ & $99.32 \pm 0.10$ \\
                                & $\ABLAVG$  & $99.60 \pm 0.09$ & $99.38 \pm 0.21$ & $99.32 \pm 0.14$ & $96.16 \pm 1.22$ & $98.46 \pm 0.33$ \\
                                & $\ABLTL$   & $99.90 \pm 0.02$ & $99.77 \pm 0.04$ & $99.21 \pm 0.00$ & $98.50 \pm 0.16$ & $99.21 \pm 0.06$ \\ \bottomrule
\end{tabular}
\end{sc}
}
}
\end{small}
\caption{Test accuracy (\%) of each method using MLP on benchmark datasets and tasks.}
\label{tab:exp-mlp}
\vspace{-4px}
\end{table*}

Corollary~\ref{coro:rank-criterion} applies directly to the task of learning with the single target concept ``$\mathtt{conj}$'' in \cref{exp:conj}.
As discussed before, the candidate labels in this task under the uniform assumption lead to the following probability matrix
$$
\textstyle
Q=
\left(\begin{array}{ccc}
2/7 & 2/7 & 3/7 \\
2/5 & 2/5 & 1/5 
\end{array}\right),
$$
which has full row rank. Therefore, according to Corollary~\ref{coro:rank-criterion}, the supervision signals from the knowledge base are sufficient to produce accurate classifiers.

Similarly, let us consider the case of learning with only the target concept ``$\mathtt{conj0}$'' in \cref{exp:conj01}. This can lead to the following probability matrix
$$
\textstyle
Q=
\left(\begin{array}{cc}
1/2 & 1/2 \\
1/2 & 1/2  
\end{array}\right),
$$
which has rank one. Thus, it may fail to recover ground-truth labels in this case. Fortunately, there is another target concept ``$\mathtt{conj1}$'' in \cref{exp:conj01}.
Leveraging both target concepts, we can derive a different probability matrix
$$
\textstyle
\wQ=
\left(\begin{array}{cccc}
1/2 & 1/2 & 0 & 0 \\
1/4 & 1/4 & 1/4 & 1/4 
\end{array}\right).
$$
This matrix has full row rank, thus facilitating the learning of accurate classifiers according to \cref{thrm:rank-criterion}.

\section{Experiments}

In this section, we conduct comprehensive experiments to validate the utility of the proposed criterion on various tasks. 
The code is available for download.\footnote{\url{https://github.com/AbductiveLearning/ABL-TL}}
%

\subsubsection{Tasks.}
We first examine four benchmark tasks: $\ConjEq$, $\HED$, $\Conjunction$, and $\Addition$.
The $\ConjEq$ task is a variant of the $\HED$ (i.e., handwritten equation decipherment) task adopted in \citet{dai2019bridging, huang2021fast}. The knowledge base in this task has been illustrated in \cref{fig:motivating-example}, and it accepts triplets of handwritten Boolean symbols as inputs. In contrast, the original $\HED$ task exploits the knowledge of the correctness of binary additive equations and accepts the handwritten equations composed of digits, the plus sign, and the equal sign as inputs \cite{dai2019bridging}.
Similarly, the $\Conjunction$ task is a variant of the $\Addition$ task introduced in \citet{manhaeve2018deepproblog}. The knowledge base in this task has been provided in \cref{fig:multi-concepts-example}, and it accepts pairs of handwritten Boolean symbols as inputs. In contrast, the $\Addition$ task works with a knowledge base that defines the sum of two summands, and accepts pairs of handwritten decimal digits as inputs.
Following previous work \cite{huang2021fast, cai2021abductive}, we collect training sequences for the tasks by representing the handwritten symbols using instances from benchmark datasets including \textsc{MNIST} \cite{lecun1998gradient}, \textsc{EMNIST} \cite{cohen2017emnist}, \textsc{USPS} \cite{hull1994database}, \textsc{Kuzushiji} \cite{clanuwat2018deep}, and \textsc{Fashion} \cite{xiao2017fashion}. 
All experiments are repeated six times on GeForce RTX 3090 GPUs, with the mean accuracy and standard deviation reported. 
More details on experimental settings are found in \cref{app:experimental-settings}.

\subsubsection{Methods.}
We consider four strategies for selecting the abduced labels from the candidate set in hybrid learning systems: \textsc{Rand} (Random), \textsc{MaxP} (Maximal Probability), \textsc{MinD} (Minimal Distance), and \textsc{Avg} (Average).
Specifically, \ABLRAND~selects a consistent label sequence randomly from the candidate set for each input sequence.
\ABLMAXP~identifies the most probable labels from the candidate set, based on the likelihood $p(\bY|X)$ estimated by the classifier. This strategy aligns with common practices in previous studies \cite{li2020closed, dai2021abductive, huang2021fast}.
\ABLMIND~chooses the abduced labels that have the smallest Hamming distance to the predicted labels. This strategy also mirrors approaches found in earlier research \cite{dai2019bridging, tsamoura2021neural, cai2021abductive}.
\ABLAVG~regards all label sequences in the candidate set as plausible labels. It calculates a loss for each label sequence and then averages them. Thus, in expectation, this strategy is equivalent to the \textsc{Rand} strategy.
It is worth noting that the \textsc{MaxP} and \textsc{MinD} strategies behave similarly to the \textsc{Rand} strategy in the initial stages of training, as both the estimated probabilities and the predicted labels are essentially random when the classifier is initialised randomly.
After the above label abduction procedure, the empirical counterpart of \cref{eq:mi-risk} is used as the learning objective.
The above methods are compared with \textsc{TL}, i.e., the method of minimising the empirical counterpart of the risk in \cref{eq:tl-risk}.

\subsubsection{Experimental Results on \texttt{ConjEq} and \texttt{Conjunction}.}
As indicated by \cref{fig:motivating-example} and \cref{fig:multi-concepts-example}, both tasks aim to learn a binary classifier that perceives raw inputs and assigns them a prediction of either 0 or 1.
While the two tasks differ in terms of the number of target concepts (one versus two) and the length of input sequences (three versus two), they both satisfy the rank criterion.
Thus, it is expected that neuro-symbolic learning can produce accurate classifiers in these cases.
Indeed, this is confirmed by our experimental results.
\cref{tab:exp-mlp} presents the test performance of multi-layer perception (MLP) produced by hybrid learning methods on various datasets for the $\ConjEq$ and $\Conjunction$ tasks.
Results indicate that all four strategies for selecting abduced labels work well in facilitating successful learning: they all achieve over 98\% accuracy, and none appear to have a significant edge over the others. This implies that the \textsc{MaxP} and \textsc{MinD} strategies are empirically similar to the \textsc{Rand} strategy in the $\ConjEq$ and $\Conjunction$ tasks.
The phenomenon can be explained by the small candidate set size ($\le 4$) for label abduction in both tasks, which leads to a considerable probability of correctly selecting the labels even when employing the \textsc{Rand} strategy.
We also observe that \ABLTL~consistently achieves competitive performance across all datasets in both tasks, which corroborates our theoretical analysis.

\subsubsection{Experimental Results on \texttt{Addition}.}
This task is more challenging than $\Conjunction$, as the size of the candidate set in $\Addition$ is larger, which complicates the abduction of correct labels. This difficulty is amplified by the presence of 10 classes in $\Addition$, compared to just 2 classes in $\Conjunction$.
Despite these challenges, our rank criterion positively indicates that the supervision signals from the knowledge base of $\Addition$ are sufficient for learning accurate classifiers.
This is confirmed by \cref{tab:exp-mlp}, showing that all methods can learn classifiers with significantly greater accuracy than random guessing in the $\Addition$ task.
An insightful observation is that the \textsc{MaxP} and \textsc{MinD} strategies consistently outperform the \textsc{Rand} strategy in this task. This implies that, despite their similar behaviours in the early stages of training, these strategies may diverge in later training stages. As the accuracy of the estimated probabilities and the predicted labels improves, \textsc{MaxP} and \textsc{MinD} are increasingly likely to select the correct labels over \textsc{Rand}. Finally, we observe that \ABLTL~outperforms \ABLRAND~and \ABLAVG, while showing competitive results with \ABLMAXP~and \ABLMIND.

\subsubsection{Experimental Results on \texttt{HED}.}
Results on $\HED$ using the knowledge base of binary additive equations can be found in \cref{tab:exp-mlp}.
Again, all methods consistently perform well across all datasets for $\HED$. 
This success is also indicated by our rank criterion. The knowledge base of binary additive equations satisfies the criterion: the rank of the corresponding probability matrix is 4, which equals the number of symbols in binary additive equations (i.e., ``$0$'', ``$1$'', ``$+$'', and ``$=$''). 
However, when the task is extended to handle additive equations with decimal digits, the corresponding matrix rank becomes 7, which is less than 12, the number of symbols in decimal additive equations (i.e., ``$0$'', $\ldots$, ``$9$'', ``$+$'', and ``$=$''). In other words, the knowledge base of decimal additive equations does not satisfy the criterion. This indicates that learning would fail in this case. 

Our experiments validate this, showing that $\ABLTL$ performs poorly when using the knowledge base of decimal additive equations, with test accuracy falling below $50\%$.
Thorough experiments involving additive equations across number systems from base 2 to base 10 further confirm the utility of the proposed criterion, as shown by the performance drop when the base is larger than 5. 
Finally, we note that, although the rank criterion is negative in the case of base 10 for $\HED$, \citet{huang2021fast} have shown that effective learning is still possible with additional assumptions. Concretely, they assumed that there exists a similarity between raw inputs. While this assumption may not always hold in practice, it is interesting to theoretically analyse its usefulness in helping neuro-symbolic learning. 
We leave this as future work.

\begin{figure}[t]
    \centering
    \includegraphics[width=0.42\textwidth]{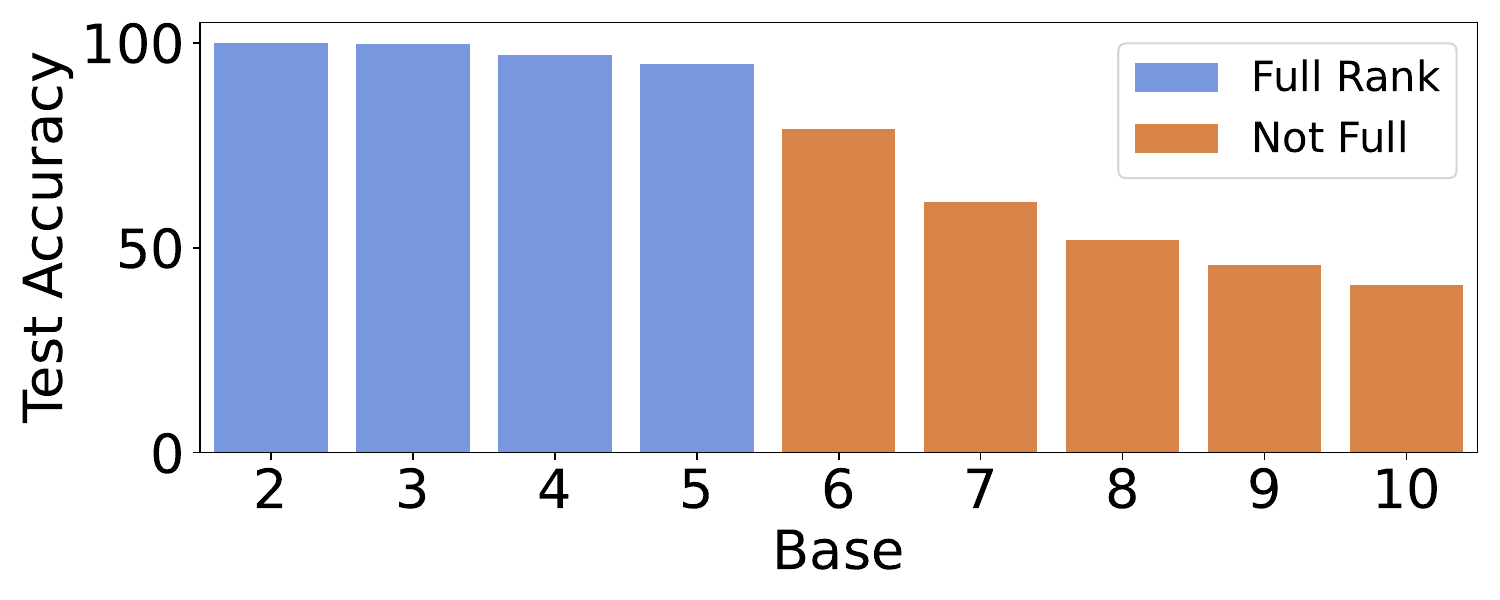}
\caption{
Performance on $\HED$ using different knowledge bases of numeral systems ranging from base 2 to base 10.
}
\label{fig:hed-resnet-tl}
\vspace{-6px}
\end{figure}

\begin{figure*}[!t]
  \centering
  \subfigure[$\DNF$, $m=3$]{
  \label{fig:random-dnf-3}
     \centering
     \includegraphics[width=0.30\textwidth]{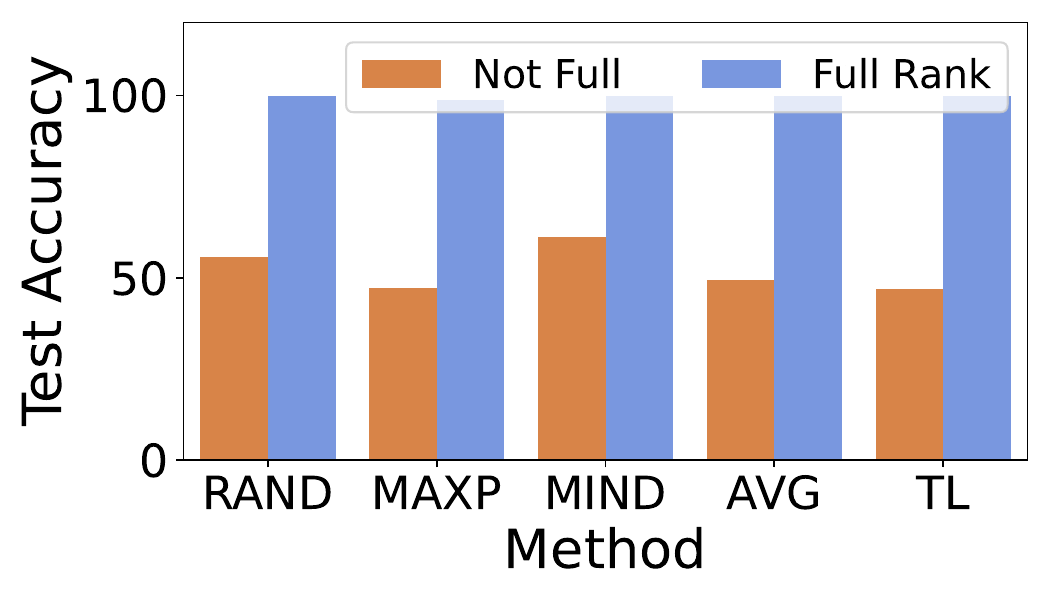}
  }
  \hfill
  \subfigure[$\DNF$, $m=4$]{
  \label{fig:random-dnf-4}
     \centering
     \includegraphics[width=0.30\textwidth]{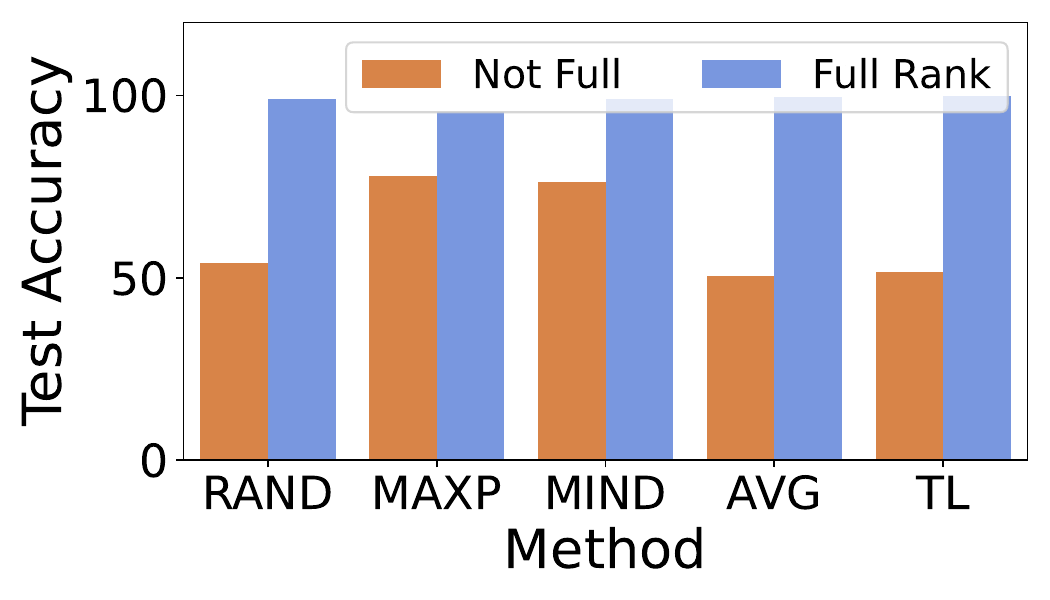}
  }
  \hfill
  \subfigure[$\DNF$, $m=5$]{
  \label{fig:random-dnf-5}
     \centering
     \includegraphics[width=0.30\textwidth]{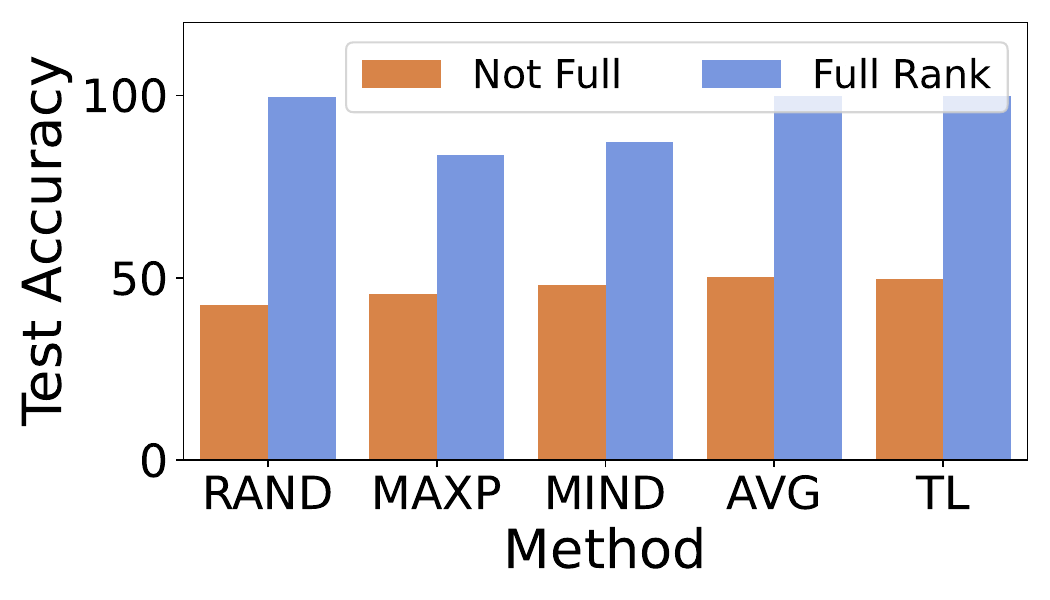}
  }
\caption{
Comparison of knowledge bases satisfying or not satisfying the rank criterion. Knowledge bases are created by randomly generating rules in disjunctive or conjunctive norm forms, with clause lengths varying from 3 to 5. The rank criterion effectively indicates the success of learning accurate classifiers.
}
\label{fig:random-knowledge-bases}
\vspace{-4px}
\end{figure*}

\subsubsection{Experimental Results on Random Knowledge Bases.}
To further demonstrate the utility of our rank criterion, we conduct experiments on random knowledge bases.
Following \citet{cai2021abductive}, we create knowledge bases by randomly generating binary rules in disjunctive normal form (DNF), i.e., disjunctions of conjunctive clauses. 
In DNF, we control the clause length $m$, and the number of clauses is randomly chosen from the range $[1, 2^{m-1}]$.
Most generated knowledge bases satisfy the rank criterion, while there are some knowledge bases whose corresponding matrix rank is deficient.
Examples of these knowledge bases are illustrated in \cref{app:random-knowledge-bases}.
Numerical results are summarised in \cref{fig:random-knowledge-bases}.
We observe that our rank criterion consistently indicates the learning performance of all methods in all cases.
When the probability matrix corresponding to a knowledge base has full row rank, the learned classifiers mostly achieve high test accuracy; conversely, rank deficiency indicates low performance, close to random guessing. 
While a few outliers exist, such as the rank-deficient case in \cref{fig:random-dnf-4}, where the \textsc{MaxP} and \textsc{MinD} strategies perform slightly better than random guessing, their results remain significantly lower than the full-rank case.
We also notice that the \textsc{MaxP} and \textsc{MinD} strategies perform slightly worse in \cref{fig:random-dnf-5}, potentially due to the size of the candidate set being large in those cases, which may complicate the abduction of correct labels. Nevertheless, we observe that \ABLRAND, \ABLAVG, and \ABLTL~consistently perform well across all cases.

\section{Related Work}

\subsubsection{Neuro-Symbolic Learning.}
Neuro-symbolic integration had been studied decades ago \cite{towell1994knowledge, garcez2002neural}.
In recent years, the increasing efficacy of modern machine learning techniques, such as deep learning, sparked a surge of interest in integrating machine learning and symbolic reasoning.
Many efforts have been devoted to this integration~\cite{dai2017combining, donadello2017logic, gaunt2017differentiable, santoro2017simple, grover2018stochastic, xu2018semantic, trask2018neural, manhaeve2018deepproblog, dai2019bridging, dong2018neural, wang2019satnet, cohen2020tensorlog, yang2021neurasp, evans2021making, li2022softened, liu2023out, huang2023enabling, huang2023enablingabductive}, with abductive learning (ABL) emerging as one of the most expressive frameworks for hybrid systems~\cite{DBLP:journals/chinaf/Zhou19, zhou2022abductive}.
It is noteworthy that, beyond the basic settings in this paper, the ABL framework is highly general and flexible, accommodating various machine learning mechanisms and allowing the exploitation of abundant labelled data and inaccurate knowledge bases \cite{DBLP:journals/chinaf/Zhou19}.
ABL has found applications in many areas, such as theft judicial sentencing~\cite{huang2020semi}, stroke evaluation~\cite{wang2021tac}, and optical character recognition~\cite{cai2021abductive}.

\subsubsection{Weakly Supervised Learning.}
The ABL framework has been viewed as enlarging the scope of weakly supervised learning (WSL) \cite{zhou2018brief}, where the supervision information can come from knowledge reasoning \cite{zhou2022abductive}. Yet, traditional WSL settings, such as multi-instance learning \cite{dietterich1997solving, zhou2009multi} and partial-label learning \cite{jin2002learning, cour2011learning, liu2014learnability, feng2020provably}, struggle to tackle the challenges in ABL. 
The consistency of learning from aggregate information has been explored in \citet{zhang2020learning} and a concurrent work by \citet{wang2023learning}, but they can only handle limited types of aggregate functions and loss functions; in contrast, our analysis applies to any type of knowledge bases under the ABL framework.
While the idea of summarising probabilities into a stochastic matrix has been widely exploited in Markov chains \cite{gagniuc2017markov} and noisy-label learning \cite{natarajan2013learning, patrini2017making, yu2018learning}, we are the first to show its relation with the objective of hybrid learning systems.
This work is also closely related to the concept of reasoning shortcuts in \citet{marconato2023neuro, marconato2023not}, where perception models may learn unintended semantics. Our rank criterion is useful in indicating such failures before actual training.

\section{Conclusion}

In this work, we introduce a novel characterisation of the supervision signals from a given knowledge base and establish a rank criterion capable of indicating the practical efficacy of a given knowledge base in improving learning performance.\\
Both theoretical and empirical results shed light on the success of hybrid learning systems while pinpointing potential failures when the supervision signals from a symbolic knowledge base are insufficient to ensure effective learning.
Future work includes the detailed analysis of mutual promotion between learning and reasoning, the incorporation of other machine learning models, and the exploitation of abundant labelled data and inaccurate knowledge bases.

\section*{Acknowledgments}
This research was supported by the NSFC (62176117, 62206124) and JiangsuSF~(BK20232003).

\bibliography{aaai24}

\begin{thebibliography}{66}
\providecommand{\natexlab}[1]{#1}

\bibitem[{Cai et~al.(2021)Cai, Dai, Huang, Li, Muggleton, and Jiang}]{cai2021abductive}
Cai, L.-W.; Dai, W.-Z.; Huang, Y.-X.; Li, Y.-F.; Muggleton, S.~H.; and Jiang, Y. 2021.
\newblock Abductive Learning with Ground Knowledge Base.
\newblock In \emph{IJCAI}, 1815--1821.

\bibitem[{Clanuwat et~al.(2018)Clanuwat, Bober-Irizar, Kitamoto, Lamb, Yamamoto, and Ha}]{clanuwat2018deep}
Clanuwat, T.; Bober-Irizar, M.; Kitamoto, A.; Lamb, A.; Yamamoto, K.; and Ha, D. 2018.
\newblock Deep learning for classical japanese literature.
\newblock \emph{arXiv preprint arXiv:1812.01718}.

\bibitem[{Cohen et~al.(2017)Cohen, Afshar, Tapson, and Van~Schaik}]{cohen2017emnist}
Cohen, G.; Afshar, S.; Tapson, J.; and Van~Schaik, A. 2017.
\newblock EMNIST: Extending MNIST to handwritten letters.
\newblock In \emph{IJCNN}, 2921--2926.

\bibitem[{Cohen, Yang, and Mazaitis(2020)}]{cohen2020tensorlog}
Cohen, W.; Yang, F.; and Mazaitis, K.~R. 2020.
\newblock Tensorlog: A probabilistic database implemented using deep-learning infrastructure.
\newblock \emph{Journal of Artificial Intelligence Research}, 67: 285--325.

\bibitem[{Cour, Sapp, and Taskar(2011)}]{cour2011learning}
Cour, T.; Sapp, B.; and Taskar, B. 2011.
\newblock Learning from partial labels.
\newblock \emph{Journal of Machine Learning Research}, 12: 1501--1536.

\bibitem[{Dai and Muggleton(2021)}]{dai2021abductive}
Dai, W.-Z.; and Muggleton, S.~H. 2021.
\newblock Abductive knowledge induction from raw data.
\newblock In \emph{IJCAI}, 1845--1851.

\bibitem[{Dai et~al.(2019)Dai, Xu, Yu, and Zhou}]{dai2019bridging}
Dai, W.-Z.; Xu, Q.; Yu, Y.; and Zhou, Z.-H. 2019.
\newblock Bridging machine learning and logical reasoning by abductive learning.
\newblock In \emph{NeurIPS}, 2811--2822.

\bibitem[{Dai and Zhou(2017)}]{dai2017combining}
Dai, W.-Z.; and Zhou, Z.-H. 2017.
\newblock Combining logical abduction and statistical induction: Discovering written primitives with human knowledge.
\newblock In \emph{AAAI}, 4392--4398.

\bibitem[{De~Raedt et~al.(2021)De~Raedt, Duman{\v{c}}i{\'c}, Manhaeve, and Marra}]{de2021statistical}
De~Raedt, L.; Duman{\v{c}}i{\'c}, S.; Manhaeve, R.; and Marra, G. 2021.
\newblock From statistical relational to neural-symbolic artificial intelligence.
\newblock In \emph{IJCAI}, 4943--4950.

\bibitem[{De~Raedt et~al.(2016)De~Raedt, Kersting, Natarajan, and Poole}]{raedt2016statistical}
De~Raedt, L.; Kersting, K.; Natarajan, S.; and Poole, D. 2016.
\newblock Statistical relational artificial intelligence: Logic, probability, and computation.
\newblock \emph{Synthesis lectures on artificial intelligence and machine learning}, 10(2): 1--189.

\bibitem[{De~Raedt and Kimmig(2015)}]{de2015probabilistic}
De~Raedt, L.; and Kimmig, A. 2015.
\newblock Probabilistic (logic) programming concepts.
\newblock \emph{Machine Learning}, 100: 5--47.

\bibitem[{Dietterich, Lathrop, and Lozano-P{\'e}rez(1997)}]{dietterich1997solving}
Dietterich, T.~G.; Lathrop, R.~H.; and Lozano-P{\'e}rez, T. 1997.
\newblock Solving the multiple instance problem with axis-parallel rectangles.
\newblock \emph{Artificial intelligence}, 89(1-2): 31--71.

\bibitem[{Donadello, Serafini, and Garcez(2017)}]{donadello2017logic}
Donadello, I.; Serafini, L.; and Garcez, A.~D. 2017.
\newblock Logic tensor networks for semantic image interpretation.
\newblock In \emph{IJCAI}.

\bibitem[{Dong et~al.(2019)Dong, Mao, Lin, Wang, Li, and Zhou}]{dong2018neural}
Dong, H.; Mao, J.; Lin, T.; Wang, C.; Li, L.; and Zhou, D. 2019.
\newblock Neural Logic Machines.
\newblock In \emph{ICLR}.

\bibitem[{Elkan and Noto(2008)}]{elkan2008learning}
Elkan, C.; and Noto, K. 2008.
\newblock Learning classifiers from only positive and unlabeled data.
\newblock In \emph{KDD}, 213--220.

\bibitem[{Evans et~al.(2021)Evans, Bo{\v{s}}njak, Buesing, Ellis, Pfau, Kohli, and Sergot}]{evans2021making}
Evans, R.; Bo{\v{s}}njak, M.; Buesing, L.; Ellis, K.; Pfau, D.; Kohli, P.; and Sergot, M. 2021.
\newblock Making sense of raw input.
\newblock \emph{Artificial Intelligence}, 299: 103521.

\bibitem[{Feng et~al.(2020)Feng, Lv, Han, Xu, Niu, Geng, An, and Sugiyama}]{feng2020provably}
Feng, L.; Lv, J.; Han, B.; Xu, M.; Niu, G.; Geng, X.; An, B.; and Sugiyama, M. 2020.
\newblock Provably consistent partial-label learning.
\newblock In \emph{NeurIPS}, 10948--10960.

\bibitem[{Gagniuc(2017)}]{gagniuc2017markov}
Gagniuc, P.~A. 2017.
\newblock \emph{Markov chains: from theory to implementation and experimentation}.
\newblock John Wiley \& Sons.

\bibitem[{Garcez, Broda, and Gabbay(2002)}]{garcez2002neural}
Garcez, A. S.~d.; Broda, K.; and Gabbay, D.~M. 2002.
\newblock \emph{Neural-symbolic learning systems: foundations and applications}.
\newblock Springer Science \& Business Media.

\bibitem[{Garcez et~al.(2007)Garcez, Gabbay, Ray, and Woods}]{garcez2007abductive}
Garcez, A. S.~d.; Gabbay, D.~M.; Ray, O.; and Woods, J. 2007.
\newblock Abductive reasoning in neural-symbolic systems.
\newblock \emph{Topoi}, 26: 37--49.

\bibitem[{Gaunt et~al.(2017)Gaunt, Brockschmidt, Kushman, and Tarlow}]{gaunt2017differentiable}
Gaunt, A.~L.; Brockschmidt, M.; Kushman, N.; and Tarlow, D. 2017.
\newblock Differentiable programs with neural libraries.
\newblock In \emph{ICML}, 1213--1222.

\bibitem[{Getoor and Taskar(2007)}]{getoor2007introduction}
Getoor, L.; and Taskar, B. 2007.
\newblock \emph{Introduction to statistical relational learning}.
\newblock MIT press.

\bibitem[{Grover et~al.(2018)Grover, Wang, Zweig, and Ermon}]{grover2018stochastic}
Grover, A.; Wang, E.; Zweig, A.; and Ermon, S. 2018.
\newblock Stochastic Optimization of Sorting Networks via Continuous Relaxations.
\newblock In \emph{ICLR}.

\bibitem[{He et~al.(2016)He, Zhang, Ren, and Sun}]{he2016deep}
He, K.; Zhang, X.; Ren, S.; and Sun, J. 2016.
\newblock Deep residual learning for image recognition.
\newblock In \emph{CVPR}, 770--778.

\bibitem[{Hitzler and Sarker(2022)}]{hitzler2022neuro}
Hitzler, P.; and Sarker, M.~K. 2022.
\newblock \emph{Neuro-Symbolic Artificial Intelligence - The State of the Art}.
\newblock Frontiers in Artificial Intelligence and Applications. IOS Press.

\bibitem[{Huang et~al.(2021)Huang, Dai, Cai, Muggleton, and Jiang}]{huang2021fast}
Huang, Y.-X.; Dai, W.-Z.; Cai, L.-W.; Muggleton, S.~H.; and Jiang, Y. 2021.
\newblock Fast abductive learning by similarity-based consistency optimization.
\newblock In \emph{NeurIPS}, 26574--26584.

\bibitem[{Huang et~al.(2023{\natexlab{a}})Huang, Dai, Jiang, and Zhou}]{huang2023enabling}
Huang, Y.-X.; Dai, W.-Z.; Jiang, Y.; and Zhou, Z.-H. 2023{\natexlab{a}}.
\newblock Enabling Knowledge Refinement upon New Concepts in Abductive Learning.
\newblock In \emph{AAAI}, 7928--7935.

\bibitem[{Huang et~al.(2020)Huang, Dai, Yang, Cai, Cheng, Huang, Li, and Zhou}]{huang2020semi}
Huang, Y.-X.; Dai, W.-Z.; Yang, J.; Cai, L.-W.; Cheng, S.; Huang, R.; Li, Y.-F.; and Zhou, Z.-H. 2020.
\newblock Semi-supervised abductive learning and its application to theft judicial sentencing.
\newblock In \emph{ICDM}, 1070--1075.

\bibitem[{Huang et~al.(2023{\natexlab{b}})Huang, Sun, Li, Tian, Dai, Hu, Jiang, and Zhou}]{huang2023enablingabductive}
Huang, Y.-X.; Sun, Z.; Li, G.; Tian, X.; Dai, W.-Z.; Hu, W.; Jiang, Y.; and Zhou, Z.-H. 2023{\natexlab{b}}.
\newblock Enabling abductive learning to exploit knowledge graph.
\newblock In \emph{IJCAI}, 3839--3847.

\bibitem[{Hull(1994)}]{hull1994database}
Hull, J.~J. 1994.
\newblock A database for handwritten text recognition research.
\newblock \emph{IEEE Transactions on pattern analysis and machine intelligence}, 16(5): 550--554.

\bibitem[{Jin and Ghahramani(2002)}]{jin2002learning}
Jin, R.; and Ghahramani, Z. 2002.
\newblock Learning with multiple labels.
\newblock In \emph{NeurIPS}, 897--904.

\bibitem[{Kakas, Kowalski, and Toni(1992)}]{kakas1992abductive}
Kakas, A.~C.; Kowalski, R.~A.; and Toni, F. 1992.
\newblock Abductive logic programming.
\newblock \emph{Journal of logic and computation}, 2(6): 719--770.

\bibitem[{Kingma and Ba(2015)}]{kingma2014adam}
Kingma, D.~P.; and Ba, J. 2015.
\newblock Adam: A method for stochastic optimization.
\newblock In \emph{ICLR}.

\bibitem[{LeCun et~al.(1998)LeCun, Bottou, Bengio, and Haffner}]{lecun1998gradient}
LeCun, Y.; Bottou, L.; Bengio, Y.; and Haffner, P. 1998.
\newblock Gradient-based learning applied to document recognition.
\newblock \emph{Proceedings of the IEEE}, 86(11): 2278--2324.

\bibitem[{Li et~al.(2020)Li, Huang, Hong, Chen, Wu, and Zhu}]{li2020closed}
Li, Q.; Huang, S.; Hong, Y.; Chen, Y.; Wu, Y.~N.; and Zhu, S.-C. 2020.
\newblock Closed loop neural-symbolic learning via integrating neural perception, grammar parsing, and symbolic reasoning.
\newblock In \emph{ICML}, 5884--5894.

\bibitem[{Li et~al.(2023)Li, Yao, Chen, Xu, Cao, Ma, Jian et~al.}]{li2022softened}
Li, Z.; Yao, Y.; Chen, T.; Xu, J.; Cao, C.; Ma, X.; Jian, L.; et~al. 2023.
\newblock Softened Symbol Grounding for Neuro-symbolic Systems.
\newblock In \emph{ICLR}.

\bibitem[{Little and Rubin(1987)}]{little1987statistical}
Little, R.; and Rubin, D. 1987.
\newblock \emph{Statistical Analysis With Missing Data}.
\newblock Wiley Series in Probability and Statistics. Wiley.
\newblock ISBN 9780471802549.

\bibitem[{Liu et~al.(2023)Liu, Xu, Van~den Broeck, and Liang}]{liu2023out}
Liu, A.; Xu, H.; Van~den Broeck, G.; and Liang, Y. 2023.
\newblock Out-of-Distribution Generalization by Neural-Symbolic Joint Training.
\newblock In \emph{AAAI}, 12252--12259.

\bibitem[{Liu and Dietterich(2014)}]{liu2014learnability}
Liu, L.; and Dietterich, T. 2014.
\newblock Learnability of the superset label learning problem.
\newblock In \emph{ICML}, 1629--1637.

\bibitem[{Manhaeve et~al.(2018)Manhaeve, Dumancic, Kimmig, Demeester, and De~Raedt}]{manhaeve2018deepproblog}
Manhaeve, R.; Dumancic, S.; Kimmig, A.; Demeester, T.; and De~Raedt, L. 2018.
\newblock Deepproblog: Neural probabilistic logic programming.
\newblock In \emph{NeurIPS}, 3753--3763.

\bibitem[{Marconato et~al.(2023{\natexlab{a}})Marconato, Bontempo, Ficarra, Calderara, Passerini, and Teso}]{marconato2023neuro}
Marconato, E.; Bontempo, G.; Ficarra, E.; Calderara, S.; Passerini, A.; and Teso, S. 2023{\natexlab{a}}.
\newblock {Neuro Symbolic Continual Learning: Knowledge, Reasoning Shortcuts and Concept Rehearsal}.
\newblock In \emph{ICML}, 23915--23936.

\bibitem[{Marconato et~al.(2023{\natexlab{b}})Marconato, Teso, Vergari, and Passerini}]{marconato2023not}
Marconato, E.; Teso, S.; Vergari, A.; and Passerini, A. 2023{\natexlab{b}}.
\newblock Not All Neuro-Symbolic Concepts Are Created Equal: Analysis and Mitigation of Reasoning Shortcuts.
\newblock In \emph{NeurIPS}.

\bibitem[{Muggleton(2023)}]{muggleton2023hypothesizing}
Muggleton, S.~H. 2023.
\newblock Hypothesizing an algorithm from one example: the role of specificity.
\newblock \emph{Philosophical Transactions of the Royal Society A}, 381(2251): 20220046.

\bibitem[{Natarajan et~al.(2013)Natarajan, Dhillon, Ravikumar, and Tewari}]{natarajan2013learning}
Natarajan, N.; Dhillon, I.~S.; Ravikumar, P.~K.; and Tewari, A. 2013.
\newblock Learning with noisy labels.
\newblock In \emph{NeurIPS}, 1196--1204.

\bibitem[{Paszke et~al.(2019)Paszke, Gross, Massa, Lerer, Bradbury, Chanan, Killeen, Lin, Gimelshein, Antiga et~al.}]{paszke2019pytorch}
Paszke, A.; Gross, S.; Massa, F.; Lerer, A.; Bradbury, J.; Chanan, G.; Killeen, T.; Lin, Z.; Gimelshein, N.; Antiga, L.; et~al. 2019.
\newblock Pytorch: An imperative style, high-performance deep learning library.
\newblock In \emph{NeurIPS}, 8024--8035.

\bibitem[{Patrini et~al.(2017)Patrini, Rozza, Krishna~Menon, Nock, and Qu}]{patrini2017making}
Patrini, G.; Rozza, A.; Krishna~Menon, A.; Nock, R.; and Qu, L. 2017.
\newblock Making deep neural networks robust to label noise: A loss correction approach.
\newblock In \emph{CVPR}, 1944--1952.

\bibitem[{Peirce(1955)}]{sanders1955abduction}
Peirce, C.~S. 1955.
\newblock Abduction and induction.
\newblock \emph{Philosophical Writings of Pierce}, 150--56.

\bibitem[{Russell(2015)}]{russell2015unifying}
Russell, S. 2015.
\newblock Unifying logic and probability.
\newblock \emph{Communications of the ACM}, 58(7): 88--97.

\bibitem[{Santoro et~al.(2017)Santoro, Raposo, Barrett, Malinowski, Pascanu, Battaglia, and Lillicrap}]{santoro2017simple}
Santoro, A.; Raposo, D.; Barrett, D.~G.; Malinowski, M.; Pascanu, R.; Battaglia, P.; and Lillicrap, T. 2017.
\newblock A simple neural network module for relational reasoning.
\newblock In \emph{NeurIPS}, 4967--4976.

\bibitem[{Simon and Newell(1971)}]{simon1971human}
Simon, H.~A.; and Newell, A. 1971.
\newblock Human problem solving: The state of the theory in 1970.
\newblock \emph{American psychologist}, 26(2): 145.

\bibitem[{Thoma(2017)}]{thoma2017hasyv2}
Thoma, M. 2017.
\newblock The hasyv2 dataset.
\newblock \emph{arXiv preprint arXiv:1701.08380}.

\bibitem[{Towell and Shavlik(1994)}]{towell1994knowledge}
Towell, G.~G.; and Shavlik, J.~W. 1994.
\newblock Knowledge-based artificial neural networks.
\newblock \emph{Artificial intelligence}, 70(1-2): 119--165.

\bibitem[{Trask et~al.(2018)Trask, Hill, Reed, Rae, Dyer, and Blunsom}]{trask2018neural}
Trask, A.; Hill, F.; Reed, S.~E.; Rae, J.; Dyer, C.; and Blunsom, P. 2018.
\newblock Neural arithmetic logic units.
\newblock In \emph{NeurIPS}, 8046--–8055.

\bibitem[{Tsamoura, Hospedales, and Michael(2021)}]{tsamoura2021neural}
Tsamoura, E.; Hospedales, T.; and Michael, L. 2021.
\newblock Neural-symbolic integration: A compositional perspective.
\newblock In \emph{AAAI}, 5051--5060.

\bibitem[{Wang et~al.(2021)Wang, Deng, Xie, Shu, Huang, Cai, Zhang, Zhang, Zhou, and Wu}]{wang2021tac}
Wang, J.; Deng, D.; Xie, X.; Shu, X.; Huang, Y.-X.; Cai, L.-W.; Zhang, H.; Zhang, M.-L.; Zhou, Z.-H.; and Wu, Y. 2021.
\newblock Tac-valuer: Knowledge-based stroke evaluation in table tennis.
\newblock In \emph{KDD}, 3688--3696.

\bibitem[{Wang, Tsamoura, and Roth(2023)}]{wang2023learning}
Wang, K.; Tsamoura, E.; and Roth, D. 2023.
\newblock On Learning Latent Models with Multi-Instance Weak Supervision.
\newblock In \emph{NeurIPS}.

\bibitem[{Wang et~al.(2019)Wang, Donti, Wilder, and Kolter}]{wang2019satnet}
Wang, P.-W.; Donti, P.; Wilder, B.; and Kolter, Z. 2019.
\newblock Satnet: Bridging deep learning and logical reasoning using a differentiable satisfiability solver.
\newblock In \emph{ICML}, 6545--6554.

\bibitem[{Xiao, Rasul, and Vollgraf(2017)}]{xiao2017fashion}
Xiao, H.; Rasul, K.; and Vollgraf, R. 2017.
\newblock Fashion-mnist: a novel image dataset for benchmarking machine learning algorithms.
\newblock \emph{arXiv preprint arXiv:1708.07747}.

\bibitem[{Xu et~al.(2018)Xu, Zhang, Friedman, Liang, and Broeck}]{xu2018semantic}
Xu, J.; Zhang, Z.; Friedman, T.; Liang, Y.; and Broeck, G. 2018.
\newblock A semantic loss function for deep learning with symbolic knowledge.
\newblock In \emph{ICML}, 5502--5511.

\bibitem[{Yang, Ishay, and Lee(2021)}]{yang2021neurasp}
Yang, Z.; Ishay, A.; and Lee, J. 2021.
\newblock NeurASP: embracing neural networks into answer set programming.
\newblock In \emph{IJCAI}, 1755--1762.

\bibitem[{Yu et~al.(2018)Yu, Liu, Gong, and Tao}]{yu2018learning}
Yu, X.; Liu, T.; Gong, M.; and Tao, D. 2018.
\newblock Learning with biased complementary labels.
\newblock In \emph{ECCV}, 68--83.

\bibitem[{Zhang et~al.(2020)Zhang, Charoenphakdee, Wu, and Sugiyama}]{zhang2020learning}
Zhang, Y.; Charoenphakdee, N.; Wu, Z.; and Sugiyama, M. 2020.
\newblock Learning from aggregate observations.
\newblock In \emph{NeurIPS}, 7993--8005.

\bibitem[{Zhou(2019)}]{DBLP:journals/chinaf/Zhou19}
Zhou, Z. 2019.
\newblock Abductive learning: towards bridging machine learning and logical reasoning.
\newblock \emph{Science China Information Sciences}, 62(7): 76101:1--76101:3.

\bibitem[{Zhou(2018)}]{zhou2018brief}
Zhou, Z.-H. 2018.
\newblock A brief introduction to weakly supervised learning.
\newblock \emph{National science review}, 5(1): 44--53.

\bibitem[{Zhou and Huang(2022)}]{zhou2022abductive}
Zhou, Z.-H.; and Huang, Y.-X. 2022.
\newblock Abductive Learning.
\newblock In \emph{Neuro-Symbolic Artificial Intelligence: The State of the Art}, 353--369. IOS Press.

\bibitem[{Zhou, Sun, and Li(2009)}]{zhou2009multi}
Zhou, Z.-H.; Sun, Y.-Y.; and Li, Y.-F. 2009.
\newblock Multi-instance learning by treating instances as non-iid samples.
\newblock In \emph{ICML}, 1249--1256.

\end{thebibliography}


\newpage

\appendix
\onecolumn

\begin{center}{\bf {\LARGE Supplementary Material:}}
\end{center}
\begin{center}{\bf {\Large Deciphering Raw Data in Neuro-Symbolic Learning with Provable Guarantees}}
\end{center}
\vspace{0.1in}

\section{Pseudocode for Neuro-Symbolic Learning}
\label{app:algorithms}

\begin{algorithm}[!h]
\caption{Inconsistency Minimisation with Abductive Reasoning}
\label{alg:nesy}
\SetAlgoLined
    \KwIn{Perception model $h$; Knowledge base $B$; Unlabelled data $\boldsymbol{X} = \{X^{(i)}\}_{i=0}^{\ne-1} = \{[x_0^{(i)}, x_1^{(i)}, \ldots, x_{\ni-1}^{(i)}]\}_{i=0}^{\ne-1}$; Target concepts $\boldsymbol{\tau} = \{\tau^{(i)}\}_{i=0}^{\ne-1}$; Training epoch $E$}
    \KwOut{Perception model $h$}
    \Parameter{Abduction strategy $A$}
    \For{$e \gets 0$ \KwTo $E-1$}{
        \For{$i \gets 0$ \KwTo $\ne-1$}{
            $\langle X, t \rangle \gets \langle X^{(i)}, \tau^{(i)} \rangle$ \;
            \tcp{Softmax and hardmax of perception model}
            \For{$k \gets 0$ \KwTo $\ni-1$}{
                \For{$j \gets 0$ \KwTo $\nc-1$}{
                    $\hat{p}(y=j|x=x_k) \gets \exp(h_j(x_k)) / \sum_{v=0}^{\nc-1} \exp(h_v(x_k))$ \;
                }
                $f(x_k) \gets \arg\max_{j\in[\nc]} \hat{p}(y=j|x=x_k)$ \;
            }
            $\hp(Y|X) \gets \prod_{k=0}^{\ni-1} \hat{p}(y=y_k|x=x_k)$ \tcp*{Likelihood}
            $\hat{Y} \gets [f(x_0), f(x_1), \ldots, f(x_{\ni-1})]$ \tcp*{Perceived labels} 
            $\cS(t) \gets \{Y \in \cY^\ni \mid B \cup Y \models t\}$ \tcp*{Candidate set for abduction}

            \tcp{Selection of abduced labels from the candidate set}
            \uIf{$A =$ \textsc{MaxP}}{
            \tcp{Maximal probability}
            $\bY \gets \arg\max_{Y \in \cS(t)} \hp(Y|X)$ \;
            }
            \uElseIf{$A =$ \textsc{MinD}}{
            \tcp{Minimal distance}
            $\bY \gets \arg\min_{Y \in \cS(t)} \|Y - \hat{Y}\|$ \;
            }
            \Else{
            \tcp{Random}
            $\bY \gets$ Randomly select an element $Y$ from the candidate set $\cS(t)$ \;
            }
            $\bY^{(i)} \gets \bY$ \;
        }
        $\boldsymbol{\bY} = \{\bY^{(i)}\}_{i=0}^{\ne-1} = \{[\by_0^{(i)}, \by_1^{(i)}, \ldots, \by_{\ni-1}^{(i)}]\}_{i=0}^{\ne-1}$ \;
        $h \gets \arg\min_{h\in\mathcal{H}} \frac{1}{\ne} \sum_{i=0}^{\ne-1} \cL(X^{(i)}, \bY^{(i)}; h)$ \tcp*{Update model with $\boldsymbol{X}$ and $\boldsymbol{\bY}$}
    }
\end{algorithm}

\begin{algorithm}[!h]
\caption{TL-Risk Minimisation}
\label{alg:tl-risk}
\SetAlgoLined
    \KwIn{Perception model $h$; Knowledge base $B$;Unlabelled data $\boldsymbol{X} = \{X^{(i)}\}_{i=0}^{\ne-1} = \{[x_0^{(i)}, x_1^{(i)}, \ldots, x_{\ni-1}^{(i)}]\}_{i=0}^{\ne-1}$; Target concepts $\boldsymbol{\tau} = \{\tau^{(i)}\}_{i=0}^{\ne-1}$; Training epoch $E$}
    \KwOut{Perception model $h$}
    $\wc \gets \ni \cdot |\cT|$ \;
    $\wQ \gets$ Initialise a matrix $\wQ \in \bbR^{\nc \times \wc}$\;
    \tcp{Calculation of probability matrix}
    \For{$j \gets 0$ \KwTo $\nc-1$}{
        \For{$t \gets 0$ \KwTo $|\cT|-1$}{
            $\cS(t) \gets \{Y \in \cY^\ni \mid B \cup Y \models t\}$ \;
            \For{$k \gets 0$ \KwTo $\ni-1$}{
                $p(y=j|\tau=t,\iota=k) = \sum_{Y \in \cS(t)} \ind(\by_{k} = j) p(Y) / p(t)$ \;
                $o \gets t\cdot\ni + k$ \;
                $\wQ_{jo} \gets p(y=j|\tau=t,\iota=k) p(\tau=t) p(\iota=k) / p(y=j)$ \;
            }
        }
    }
    \For{$e \gets 0$ \KwTo $E-1$}{
        \For{$i \gets 0$ \KwTo $\ne-1$}{
            $\langle X, t \rangle \gets \langle X^{(i)}, \tau^{(i)} \rangle$ \;
            \For{$k \gets 0$ \KwTo $\ni-1$}{
                $i' \gets i \cdot m + k$ \;
                $\langle x^{(i')}, \wy^{(i')} \rangle \gets \langle x_k, t \cdot \ni + k \rangle$ \tcp*{Instance-label pairs}
                \For{$j \gets 0$ \KwTo $\nc-1$}{
                    $\hat{p}(y=j|x^{(i')}) \gets \exp(h_j(x^{(i')})) / \sum_{v=0}^{\nc-1} \exp(h_v(x^{(i')}))$ \;
                }
                $g(x^{(i')}) \gets [\hp(y=0|x^{(i')}), \hp(y=1|x^{(i')}), \ldots, \hp(y=\nc-1|x^{(i')})]$ \;
                $\wq(x^{(i')}) \gets \wQ^{\top} g(x^{(i')})$ \;
            }
        }
        $h \gets \arg\min_{h\in\mathcal{H}} \frac{1}{\ne\ni} \sum_{i'=0}^{\ne\ni-1} \ell(\wq(x^{(i')}), \wy^{(i')})$ \tcp*{Update model with instance-label pairs}
    }
\end{algorithm}

\newpage

\section{Proofs}
\label{app:proofs}

\subsection{Proof of \cref{thrm:simple-upper-bound}}
\label{app:proof-simple-upper-bound}

Recall that the objective of minimal inconsistency is expressed as follows.
\begin{gather}
    \cR_{\ABL}(h) = \bbE_{p(X, \tau)} \left[ \cL(h(X), \bY) \right],
\end{gather}
where $\bY=[\by_0, \by_1, \ldots, \by_{\ni-1}]$ denotes the labels abduced from the candidate set $\cS(\tau) = \{ \bY \in \cY^\ni \mid B \cup \bY \models \tau \}$.
Although various heuristics have been proposed to select the most likely labels from the candidate set, we note that these heuristics behave like random guessing in the early stages of training when the classifier is randomly initialised.
Therefore, the case that the abduced labels are randomly chosen from $\cS(\tau)$ is especially significant.

In this case, the objective of minimal inconsistency is equivalent to the following in expectation:
\begin{equation}
\begin{aligned}
    \cR_{\ABL}(h) &= \bbE_{p(X, \tau)} \bbE_{p(Y)} \left[ \cL(h(X), Y) \right] \\
    &= \bbE_{p(X, \tau)} \left[ \frac{1}{|\cS(\tau)|} \sum_{\bY \in \cS(\tau)}  \cL(h(X), \bY) \right], \\
\end{aligned}
\end{equation}
where $p(Y) = 1/|\cS(\tau)|$ since the concept space $\cT$ contains only one target concept $\tau$ and the uniform assumption holds.

Fix an example $X \in \cX^\ni$ with $p(X) > 0$ and define $\bbE_{p(\tau | X)}[\cdot ]$ as the expectation with respect to $p(\tau | X)$. Then, we obtain
\begin{equation}
\begin{aligned}
    \bbE_{p(\tau | X)} \bbE_{p(Y)} \left[ \cL(h(X), Y) \right]
    &= \bbE_{p(\tau | X)} \left[ \frac{1}{|\cS(\tau)|} \sum_{\bY \in \cS(\tau)}  \cL(h(X), \bY) \right] \\
    &= \sum_{\tau \in \cT} \frac{p(\tau | X) }{|\cS(\tau)|} \sum_{\bY \in \cS(\tau)}  \cL(h(X), \bY) \\
    &= \frac{1}{|\cS(\tau)|} \sum_{\bY \in \cS(\tau)}  \cL(h(X), \bY),
\end{aligned}
\end{equation}
where the last equality holds because $p(\tau|X) = 1$, i.e., any $X$ with $p(X) > 0$ belongs to the target concept $\tau$.
Denote the candidate set of abduced labels as $\cS(\tau) = \{ \bY \in \cY^\ni \mid B \cup \bY \models \tau \} = \{\bY^{\langle j \rangle}\}_{j=0}^{r-1} = \{[\by^{\langle j \rangle}_0, \by^{\langle j \rangle}_1, \ldots, \by^{\langle j \rangle}_{\ni-1}]\}_{j=0}^{r-1}$, where $r = |\cS(\tau)|$. Then, we have
\begin{equation}
\begin{aligned}
\label{eq:simple-fix-example}
    \bbE_{p(\tau | X)} \bbE_{p(Y)} \left[ \cL(h(X), Y) \right]
    &= \frac{1}{|\cS(\tau)|} \sum_{\bY \in \cS(\tau)}  \cL(h(X), \bY) \\
    &= \frac{1}{r} \sum_{j=0}^{r-1}  \frac{1}{\ni} \sum_{k=0}^{\ni-1}  \ell(h(x_{k}), \by_{k}^{\langle j \rangle}) \\
    &= \frac{1}{r} \sum_{j=0}^{r-1}  \frac{1}{\ni} \sum_{k=0}^{\ni-1}  -\log \hp(\by_{k}^{\langle j \rangle} | x_{k}), \\
    &= \frac{1}{\ni} \sum_{k=0}^{\ni-1} \frac{1}{r} \sum_{j=0}^{r-1}   -\log \hp(\by_{k}^{\langle j \rangle} | x_{k}), \\
\end{aligned}
\end{equation}
where $\hp(\by | x) =  \exp({h_{\by}(x)}) / \sum_{i=0}^{\nc-1} \exp(h_i(x))$ denotes the estimation of $p(y=\by | x)$ by the classifier $h$. 

By Jensen's inequality, we have 
\begin{equation}
\begin{aligned}
\label{eq:simple-jensen-bound}
    \frac{1}{r} \sum_{j=0}^{r-1}   -\log \hp(\by_{k}^{\langle j \rangle} | x_{k})
    &\ge   -\log  \left( \frac{1}{r} \sum_{j=0}^{r-1} \hp(\by_{k}^{\langle j \rangle} | x_{k}) \right) \\
    &= \log r  -  \log \sum_{j=0}^{r-1} \hp(\by_{k}^{\langle j \rangle} | x_{k}) 
\end{aligned}
\end{equation}

Since $\by_{k}^{\langle j \rangle} \in [\nc]$ for any $k\in[\ni]$, we obtain
\begin{equation}
\label{eq:simple-transition}
\begin{aligned}
    \sum_{j=0}^{r-1} \hp(\by_{k}^{\langle j \rangle} | x_{k}) = \sum_{i=0}^{\nc-1} \sum_{j=0}^{r-1} \ind(\by_{k}^{\langle j \rangle} = i) \hp(i | x_{k}) = r \sum_{i=0}^{\nc-1} \frac{\sum_{j=0}^{r-1} \ind(\by_{k}^{\langle j \rangle} = i)}{r} \hp(i | x_{k}),
\end{aligned}
\end{equation}
where $\ind(\cdot)$ is the indicator function.
By the uniform assumption, i.e., $p(\bY) = p([\by_0, \ldots, \by_{\ni-1}]) = 1/|\cS(\tau)|$, $\forall \bY \in \cS(\tau)$, we obtain $p(y=i \mid \iota=k) = p(y_{k}=i) = \sum_{\bY \in \cS(\tau)} \ind(\by_{k} = i) / |\cS(\tau)|$, $\forall i\in [\nc], k\in [\ni]$, and $p(y=i) = \sum_{k=0}^{\ni-1} p(y=i \mid \iota=k) p(\iota=k) = \sum_{\bY\in\cS(\tau)} \sum_{k=0}^{\ni-1} \ind(\by_{k}=i) / (rm)$, $\forall i\in [\nc]$.
Meanwhile, we have
\begin{equation}
\begin{aligned}
    p(y=i \mid \iota=k) = \frac{p(y=i)}{p(\iota=k)} \cdot p(\iota=k \mid y=i) \le a \cdot p(\iota=k \mid y=i),
\end{aligned}
\end{equation}
where the inequality holds because $a = m \cdot \max_{i\in\cY}p(y=i)$ and $p(\iota=k)=1/\ni, \forall k \in [\ni]$. 
Hence, combining this inequality with \cref{eq:simple-transition} yields
\begin{equation}
\begin{aligned}
    \sum_{j=0}^{r-1} \hp(\by_{k}^{\langle j \rangle} | x_{k}) = r \sum_{i=0}^{\nc-1} p(y=i | \iota=k) \hp(i | x_{k}) \le ar \sum_{i=0}^{\nc-1} p(\iota=k | y=i) \hp(i | x_{k}).
\end{aligned}
\end{equation}
Further, by combining this inequality with \cref{eq:simple-jensen-bound}, we obtain
\begin{equation}
\begin{aligned}
    \frac{1}{r} \sum_{j=0}^{r-1}   -\log \hp(\by_{k}^{\langle j \rangle} | x_{k})
    &\ge \log r  -  \log \sum_{j=0}^{r-1} \hp(\by_{k}^{\langle j \rangle} | x_{k}) \\
    &\ge \log r - \log \left( ar \sum_{i=0}^{\nc-1} p(\iota=k | y=i) \hp(i | x_{k}) \right) \\
    &= - \log a - \log \sum_{i=0}^{\nc-1} p(\iota=k | y=i) \hp(i | x_{k}), \\
\end{aligned}
\end{equation}
and then 
\begin{equation}
\begin{aligned}
    \frac{1}{\ni} \sum_{k=0}^{\ni-1} \frac{1}{r} \sum_{j=0}^{r-1}   -\log \hp(\by_{k}^{\langle j \rangle} | x_{k})
    &\ge - \log a - \frac{1}{\ni} \sum_{k=0}^{\ni-1} \log \sum_{i=0}^{\nc-1} p(\iota=k | y=i) \hp(i | x_{k}) \\
    &= - \log a - \frac{1}{\ni} \sum_{k=0}^{\ni-1} \log \sum_{i=0}^{\nc-1} Q_{ik} \hp(i | x_{k}) \\
    &= - \log a - \frac{1}{\ni} \sum_{k=0}^{\ni-1} \log q_k(x_{k}) \\
    &= - \log a + \frac{1}{\ni} \sum_{k=0}^{\ni-1} \ell(q(x_{k}), k) \\
\end{aligned}
\end{equation}
where the penultimate equality holds since we have defined $q(x) = Q^\top g(x)$ as an estimation of the conditional probability $p(\iota|x)$ in \cref{eq:simple-reformulation} and $g_i(x) = \hp(i | x) = \exp({h_{i}(x)}) / \sum_{j=0}^{\nc-1} \exp(h_j(x)), \forall i \in \cY$.
Meanwhile, according to the generation process of the instance-location pairs described in \cref{sec:motivating-example}, we have
\begin{equation}
\begin{aligned}
    \cR_{\L}(h) = \bbE_{p(x, \iota)} \ell(q(x), \iota)
    = \bbE_{p(X, \tau)} \left[ \frac{1}{\ni} \sum_{k=0}^{\ni-1} \ell(q(x_{k}), k) \right].
\end{aligned}
\end{equation}

Finally, we conclude by taking expectation over $X$ in \cref{eq:simple-fix-example} as follows.
\begin{equation}
\begin{aligned}
    \cR_{\ABL}(h)
    &= \bbE_{p(X, \tau)} \left[ \frac{1}{|\cS(\tau)|} \sum_{\bY \in \cS(\tau)}  \cL(h(X), \bY) \right] \\
    &\ge \bbE_{p(X, \tau)} \left[ \frac{1}{\ni} \sum_{k=0}^{\ni-1} \ell(q(x_{k+1}), k) \right] -\log a \\
    &= \cR_{\L}(h) - \log a.
\end{aligned}
\end{equation}
\qed

\subsection{Proof of \cref{thrm:upper-bound-general-case}}
\label{app:proof-upper-bound}

Recall that the objective of minimal inconsistency is expressed as follows.
\begin{gather}
    \cR_{\ABL}(h) = \bbE_{p(X, \tau)} \left[ \cL(h(X), \bY) \right],
\end{gather}
where $\bY=[\by_0, \by_1, \ldots, \by_{\ni-1}]$ denotes the labels abduced from the candidate set $\cS(\tau) = \{ \bY \in \cY^\ni \mid B \cup \bY \models \tau \}$.
Although various heuristics have been proposed to select the most likely labels from the candidate set, we note that these heuristics behave like random guessing in the early stages of training when the classifier is randomly initialised.
Therefore, the case that the abduced labels are randomly chosen from $\cS(\tau)$ is especially significant.

In this case, the objective of minimal inconsistency is equivalent to the following in expectation:
\begin{equation}
\begin{aligned}
    \cR_{\ABL}(h) &= \bbE_{p(X, \tau)} \bbE_{p(Y|\tau)} \left[ \cL(h(X), Y) \right] \\
    &= \bbE_{p(X, \tau)} \left[ \sum_{\bY \in \cS(\tau)} p(Y=\bY | \tau) \cL(h(X), \bY) \right]. \\
\end{aligned}
\end{equation}

Fix an example $X \in \cX^\ni$ and a target concept $t \in \cT$ such that $p(X, t) > 0$. 
Denote the candidate set of abduced labels as $\cS(t) = \{ \bY \in \cY^\ni \mid B \cup \bY \models t \} = \{\bY^{\langle j \rangle}\}_{j=0}^{r-1} = \{[\by^{\langle j \rangle}_0, \by^{\langle j \rangle}_1, \ldots, \by^{\langle j \rangle}_{\ni-1}]\}_{j=0}^{r-1}$, where $r = |\cS(\tau)|$.
Then, the objective becomes
\begin{equation}
\begin{aligned}
\label{eq:general-fix-example}
    \sum_{\bY \in \cS(t)} p(Y=\bY | \tau=t) \cL(h(X), \bY)
    &= \sum_{j=0}^{r-1} p(Y=\bY^{\langle j \rangle} | \tau=t) \cL(h(X), \bY^{\langle j \rangle}) \\
    &= \sum_{j=0}^{r-1} p(Y=\bY^{\langle j \rangle} | \tau=t) \frac{1}{\ni} \sum_{k=0}^{\ni-1}  \ell(h(x_{k}), \by_{k}^{\langle j \rangle}) \\
    &= \frac{1}{\ni} \sum_{k=0}^{\ni-1} \sum_{j=0}^{r-1} p(Y=\bY^{\langle j \rangle} | \tau=t)  \ell(h(x_{k}), \by_{k}^{\langle j \rangle}) \\
    &= \frac{1}{\ni} \sum_{k=0}^{\ni-1} \sum_{j=0}^{r-1} p(Y=\bY^{\langle j \rangle} | \tau=t)  \cdot \left( -\log \hp(\by_{k}^{\langle j \rangle} | x_{k}) \right) \\
\end{aligned}
\end{equation}
where $\hp(\by | x) =  \exp({h_{\by}(x)}) / \sum_{i=0}^{\nc-1} \exp(h_i(x))$ denotes the estimation of $p(y=\by | x)$ by the classifier $h$. 

By Jensen's inequality, we have 
\begin{equation}
\begin{aligned}
\label{eq:general-jensen-bound}
    \sum_{j=0}^{r-1} p(Y=\bY^{\langle j \rangle} | \tau=t)  \cdot \left( -\log \hp(\by_{k}^{\langle j \rangle} | x_{k}) \right)
    &\ge   -\log   \sum_{j=0}^{r-1} p(Y=\bY^{\langle j \rangle} | \tau=t) \hp(\by_{k}^{\langle j \rangle} | x_{k}) \\
\end{aligned}
\end{equation}

Since $\by_{k}^{\langle j \rangle} \in [\nc]$ for any $k\in[\ni]$, we obtain
\begin{equation}
\label{eq:general-transition}
\begin{aligned}
    \sum_{j=0}^{r-1} p(Y=\bY^{\langle j \rangle} | \tau = t) \hp(\by_{k}^{\langle j \rangle} | x_{k}) 
    &= \sum_{i=0}^{\nc-1} \sum_{j=0}^{r-1} \ind(\by_{k}^{\langle j \rangle} = i) p(Y=\bY^{\langle j \rangle} | \tau = t) \hp(i | x_{k}),
\end{aligned}
\end{equation}
where $\ind(\cdot)$ is the indicator function.
By marginalising $p(y_{k}, Y | \tau)$ over $Y$, we obtain
$p(y=i \mid \iota=k, \tau=t) = p(y_{k}=i \mid \tau=t) = \sum_{\bY\in\cS(t)} p(y_{k}=i | Y=\bY) p(Y=\bY | \tau=t) = \sum_{\bY\in\cS(t)} \ind(\by_{k} = i) p(Y=\bY | \tau=t)$, $\forall i\in[\nc], k\in[\ni], t\in\cT$.
Meanwhile, we have
\begin{equation}
\begin{aligned}
    p(y=i \mid \iota=k, \tau=t) = \frac{p(y=i)}{p(\iota=k) p(\tau=t)} \cdot p(\tau=t, \iota=k \mid y=i) \le \frac{m}{b} \cdot p(\tau=t, \iota=k \mid y=i),
\end{aligned}
\end{equation}
where the inequality holds because $p(y=i) \ge 1, \forall i\in\cY$, $p(\iota=k) = 1/\ni, \forall k\in[\ni]$, and $b = \min_{t\in\cT} p(\tau=t)$.
Hence, combining this inequality with \cref{eq:general-transition} yields
\begin{equation}
\begin{aligned}
    \sum_{j=0}^{r-1} p(Y=\bY^{\langle j \rangle} | \tau = t) \hp(\by_{k}^{\langle j \rangle} | x_{k}) 
    &= \sum_{i=0}^{\nc-1} p(y=i \mid \iota=k, \tau=t) \hp(i | x_{k}) \\
    &\le \frac{\ni}{b} \sum_{i=0}^{\nc-1} p(\tau=t, \iota=k \mid y=i) \hp(i | x_{k}).
\end{aligned}
\end{equation}
Further, by combining this inequality with \cref{eq:general-jensen-bound}, we obtain
\begin{equation}
\begin{aligned}
    \sum_{j=0}^{r-1} p(Y=\bY^{\langle j \rangle} | \tau=t)  \cdot \left( -\log \hp(\by_{k}^{\langle j \rangle} | x_{k}) \right)
    &\ge   -\log   \sum_{j=0}^{r-1} p(Y=\bY^{\langle j \rangle} | \tau=t) \hp(\by_{k}^{\langle j \rangle} | x_{k}) \\
    &\ge -\log \left( \frac{\ni}{b} \sum_{i=0}^{\nc-1} p(\tau=t, \iota=k \mid y=i) \hp(i | x_{k}) \right) \\
    &= \log \frac{b}{\ni} - \log \sum_{i=0}^{\nc-1} p(\tau=t, \iota=k \mid y=i) \hp(i | x_{k}) \\
    &= \log \frac{b}{\ni} - \log \sum_{i=0}^{\nc-1} p(\wy=o \mid y=i) \hp(i | x_{k}) \\
    &= \log \frac{b}{\ni} - \log \sum_{i=0}^{\nc-1} \wQ_{io} \hp(i | x_{k}),
\end{aligned}
\end{equation}
where $\wy$ represents a synthetic label with value $o = t\cdot\ni + k$, and $\wQ_{io}$ denotes $p(\wy=o \mid y=i)$.
Then, we have
\begin{equation}
\begin{aligned}
    \frac{1}{\ni} \sum_{k=0}^{\ni-1} \sum_{j=0}^{r-1} p(Y=\bY^{\langle j \rangle} | \tau=t)  \cdot \left( -\log \hp(\by_{k}^{\langle j \rangle} | x_{k}) \right)
    &\ge \log \frac{b}{\ni} - \frac{1}{\ni} \sum_{k=0}^{\ni-1} \log \sum_{i=0}^{\nc-1} \wQ_{io} \hp(i | x_{k}) \\
    &= \log \frac{b}{\ni} - \frac{1}{\ni} \sum_{k=0}^{\ni-1} \log \wq_o(x_{k}) \\
    &= \log \frac{b}{\ni} + \frac{1}{\ni} \sum_{k=0}^{\ni-1} \ell(\wq(x_{k}), o) \\
\end{aligned}
\end{equation}
where the penultimate equality holds since we have defined $\wq(x) = \wQ^\top g(x)$ as an estimation of the conditional probability $p(\wy|x)$ in \cref{eq:general-reformulation} and $g_i(x) = \hp(i | x) = \exp({h_{i}(x)}) / \sum_{j=0}^{\nc-1} \exp(h_j(x)), \forall i \in \cY$.
Meanwhile, according to the generation process of the instance-target-location triplets described in \cref{eq:upper-bound}, we have
\begin{equation}
    \cR_{\TL}(h) = \bbE_{p(x, \wy)} \ell(\wq(x), \wy) = \bbE_{p(X, \tau)} \left[ \frac{1}{\ni} \sum_{k=0}^{\ni-1} \ell(\wq(x_{k}), o) \right].
\end{equation}

Finally, we conclude by taking expectation over $X \in \cX^\ni$ and $t\in\cT$ in \cref{eq:general-fix-example} as follows.
\begin{equation}
\begin{aligned}
    \cR_{\ABL}(h)
    &= \bbE_{p(X,\tau)} \left[ \sum_{\bY \in \cS(\tau)} p(Y=\bY | \tau) \cL(h(X), \bY) \right] \\
    &\ge \bbE_{p(X,\tau)} \left[ \frac{1}{\ni} \sum_{k=0}^{\ni-1} \ell(\wq(x_{k}), o) \right] + \log\frac{b}{\ni} \\
    &= \cR_{\TL}(h) + \log\frac{b}{\ni}.
\end{aligned}
\end{equation}
\qed

\subsection{Proof of \cref{thrm:rank-criterion}}
\label{app:proof-rank-criterion}

By definition, we have

\begin{equation}
\begin{aligned}
\cR_{\TL}(h) &= \bbE_{p(x, \wy)}  \ell(\wq(x), \wy) \\
&= \int_{x\in\cX} \left[ \sum_{i=0}^{\nc-1} \ell(\wq(x), i) p(\wy=i|x) \right] p(x) dx \\
&= \int_{x\in\cX} \left[ - \sum_{i=0}^{\nc-1} p(\wy=i|x) \log \wq_i(x) \right] p(x) dx. \\
\end{aligned}
\end{equation}

Note that when $\cR_{\TL}(h)$ is minimised, $-\sum_{i=0}^{\nc-1} p(\wy=i|x) \log \wq_i(x)$ is also minimised for any $x$ with $p(x) > 0$.
For cross-entropy loss, we have the following optimisation problem:

\begin{equation}
\begin{aligned}
\min_{\wq} & -\sum_{i=0}^{\nc-1} p(\wy=i|x) \log \wq_i(x), \\
\text{s.t.} & \sum_{i=0}^{\nc-1} \wq_i(x) = 1.
\end{aligned}
\end{equation}

By using the Lagrange multiplier method, we have

\begin{equation}
\begin{aligned}
\min_{\wq} & -\sum_{i=0}^{\nc-1} p(\wy=i|x) \log \wq_i(x) + \lambda (\sum_{i=0}^{\nc-1} \wq_i(x) - 1).
\end{aligned}
\end{equation}

By setting the derivative to $0$, we obtain $\wq_i^{*}(x) = \frac{1}{\lambda} p(\wy = i | x)$.
Meanwhile, since $\sum_{i=0}^{\nc-1} \wq_i^{*}(x) = \sum_{i=0}^{\nc-1} p(\wy = i | x) = 1$, we have $\lambda = 1$. Then, we obtain $\wq_i^{*}(x) = p(\wy = i | x)$ for any $x$ with $p(x) > 0$, i.e., $\wq^*(x) = p(\wy|x)$.

Similarly, by definition, we have

\begin{equation}
\begin{aligned}
\cR(h) &= \bbE_{p(x, y)}  \ell(h(x), y) \\
&= \int_{x\in\cX} \left[ \sum_{i=0}^{\nc-1} \ell(h(x), i) p(y=i|x) \right] p(x) dx \\
&= \int_{x\in\cX} \left[ - \sum_{i=0}^{\nc-1} p(y=i|x) \log g_i(x) \right] p(x) dx, \\
\end{aligned}
\end{equation}
where $g_{j}(x) = \exp({h_{j}(x)}) / \sum_{i=0}^{\nc-1} \exp(h_i(x))$, i.e., $g(x) = \operatorname{softmax}(h(x))$.
Then, the minimiser of $\cR(h)$ is $g^*(x) = p(y|x)$, i.e., $h^{*}(x) = \operatorname{softmax}^{-1}(g^*(x)) = \operatorname{softmax}^{-1}(p(y|x))$.

Therefore, we obtain
\begin{equation}
\begin{aligned}
\label{eq:true-minimiser}
\wq^{*}(x) = p(\wy|x) = \wQ^{\top} p(y|x) = \wQ^{\top} \operatorname{softmax}(h^*(x)).
\end{aligned}
\end{equation}

On the other hand, by the definition of $\wq(x) = \wQ^{\top} g(x) = \wQ^{\top} \operatorname{softmax}(h(x))$, the minimiser of $\cR_{\TL}(h)$, denoted by $h^*_{\TL}(x)$, satisfies the following:
\begin{equation}
\begin{aligned}
\label{eq:tl-minimiser}
\wq^{*}(x) = \wQ^{\top} \operatorname{softmax}(h^*_{\TL}(x)).
\end{aligned}
\end{equation}

By combining \cref{eq:true-minimiser} and \cref{eq:tl-minimiser}, we have
\begin{equation}
\begin{aligned}
\wQ^{\top} \operatorname{softmax}(h^*(x)) = \wQ^{\top} \operatorname{softmax}(h^*_{\TL}(x)).
\end{aligned}
\end{equation}

Therefore, if $\wQ$ has full row rank, we obtain $\operatorname{softmax}(h^*(x)) = \operatorname{softmax}(h^*_{\TL}(x))$, which implies $h^*(x) = h^*_{\TL}(x)$.
\qed

\section{Experimental Settings}
\label{app:experimental-settings}

\paragraph{Knowledge Bases.}
We considered tasks with different knowledge bases, including $\ConjEq$, $\HED$ \cite{dai2019bridging}, $\Conjunction$, $\Addition$ \cite{manhaeve2018deepproblog}, and randomly generated $\DNF$/$\CNF$ \cite{cai2021abductive}. 
Specifically,
\begin{itemize}
    \item $\ConjEq$: The knowledge base and the facts abduced from it are shown in \cref{fig:kb-conjeq,fig:kb-conjeq-facts}. The concept space is $\cT = \{\mathtt{conj}\}$, the label space is $\cY = \{0, 1\}$, and the number of instances in a sequence is $3$.
    \item $\HED$: The facts abduced from the knowledge base of binary additive equations with lengths between 5 and 7 are shown in \cref{fig:kb-hed-2-facts}, while the details of the knowledge base can be found in \citet{dai2019bridging} and \citet{huang2021fast}. The concept space is $\cT = \{\mathtt{equation5}, \mathtt{equation6}, \mathtt{equation7}\}$, the label space is $\cY = \{0, 1, +, =\}$, and the number of instances in a sequence is between 5 and 7.
    \item $\Conjunction$: The knowledge base and the facts abduced from it are shown in \cref{fig:kb-conjunction,fig:kb-conjunction-facts}. The concept space is $\cT = \{\mathtt{conj0}, \mathtt{conj1}\}$, the label space is $\cY = \{0, 1\}$, and the number of instances in a sequence is $2$.
    \item $\Addition$: The knowledge base and the facts abduced from it are shown in \cref{fig:kb-addition,fig:kb-addition-facts}. The concept space is $\cT = \{\mathtt{zero}, \mathtt{one}, \mathtt{two}, \ldots, \mathtt{eighteen}\}$, the label space is $\cY = \{0, 1, \ldots, 9\}$, and the number of instances in a sequence is $2$.
    \item $\DNF$/$\CNF$: The knowledge bases are created by randomly generating binary rules in disjunctive normal form or conjunctive normal form. The concept space contains two target concepts, the label space is $\cY = \{0, 1\}$, and the number of instances in a sequence is $3$, $4$, or $5$. Examples of random knowledge bases with $m=3$ are shown in \cref{app:random-knowledge-bases}.
\end{itemize}

\paragraph{Datasets.}
We collected training sequences under the uniform assumption by representing the handwritten digital symbols using instances from benchmark datasets including \textsc{MNIST} \cite{lecun1998gradient}, \textsc{EMNIST} \cite{cohen2017emnist}, \textsc{USPS} \cite{hull1994database}, \textsc{Kuzushiji} \cite{clanuwat2018deep}, and \textsc{Fashion} \cite{xiao2017fashion}. Additionally, the symbols of plus and equiv were collected from the \textsc{HASYv2} dataset \cite{thoma2017hasyv2}.
Specifically,
\begin{itemize}
    \item \textsc{MNIST}: It is a 10-class dataset of handwritten digits (0 to 9). Each instance is a $28\times 28$ grayscale image.
    \item \textsc{EMNIST}: It is an extension of MNIST to handwritten letters. We use it as a drop-in replacement for the original MNIST dataset. Each instance is a $28\times 28$ grayscale image.
    \item \textsc{USPS}: It is a 10-class dataset of handwritten digits (0 to 9). Each instance is a $16\times 16$ grayscale image. We resized these images into $28\times 28$
    \item \textsc{Kuzushiji}: It is a 10-class dataset of cursive Japanese characters. Each instance is a $28\times 28$ grayscale image.
    \item \textsc{Fashion}: It is a 10-class dataset of fashion items. Each instance is a $28\times 28$ grayscale image.
    \item \textsc{HASYv2}: It is a 369-class dataset of handwritten symbols. Each instance is a $32\times 32$ grayscale image. We resized these images into $28\times 28$. The ``plus'' and ``equiv'' in this dataset were collected for the $\HED$ task.
\end{itemize}

We implement the algorithms with PyTorch \cite{paszke2019pytorch} and use the Adam \cite{kingma2014adam} optimiser with a mini-batch size set to $256$ and a learning rate set to $0.001$ for $100$ epochs.
All experiments are repeated six times on GeForce RTX 3090 GPUs, with the mean accuracy and standard deviation reported.

\begin{figure*}[!h]
    \centering
    \fbox{
    \begin{minipage}{0.4\linewidth}
    \small
        \texttt{conj([Y0,Y1,Y2]) $\gets$ Y2 is Y0 $\wedge$ Y1.
        }
    \end{minipage}
    }
\caption{A knowledge base of $\ConjEq$.     }
\label{fig:kb-conjeq}
\end{figure*}

\begin{figure*}[!h]
    \centering
    \fbox{
    \begin{minipage}{0.4\linewidth}
    \small
        \texttt{conj([0,0,0]. \\
        conj([0,1,0]. \\
        conj([1,0,0]. \\
        conj([1,1,1]. 
        }
    \end{minipage}
    }
\caption{The facts abduced from the knowledge base of $\ConjEq$ in \cref{fig:kb-conjeq}.     }
\label{fig:kb-conjeq-facts}
\end{figure*}

\begin{figure*}[!h]
    \centering
    \fbox{
    \begin{minipage}{0.4\linewidth}
    \small
        \texttt{conj0([Y0,Y1]) $\gets$ 0 is Y0 $\wedge$ Y1. \\
        conj1([Y0,Y1]) $\gets$ 1 is Y0 $\wedge$ Y1. 
        }
    \end{minipage}
    } 
\caption{A knowledge base of $\Conjunction$.    }
\label{fig:kb-conjunction}
\end{figure*}

\begin{figure*}[!h]
    \centering
    \fbox{
    \begin{minipage}{0.4\linewidth}
    \small
        \texttt{conj0([0,0]. \\
        conj0([0,1]. \\
        conj0([1,0]. \\
        conj1([1,1]. 
        }
    \end{minipage}
    }
\caption{Facts abduced from the knowledge base of $\Conjunction$ in \cref{fig:kb-conjunction}.     }
\label{fig:kb-conjunction-facts}
\end{figure*}

\begin{figure*}[!h]
    \centering
    \fbox{
    \begin{minipage}{0.4\linewidth}
    \small
        \texttt{equation5([0,+,0,=,0]). \\
        equation5([0,+,1,=,1]). \\
        equation5([1,+,0,=,1]). \\
        equation6([1,+,1,=,1,0]). \\
        equation7([0,+,1,0,=,1,0]). \\
        equation7([0,+,1,1,=,1,1]). \\
        equation7([1,0,+,0,=,1,0]). \\
        equation7([1,0,+,1,=,1,1]). \\
        equation7([1,1,+,0,=,1,1]). \\
        equation7([1,+,1,0,=,1,1]).
        }
    \end{minipage}
    }
\caption{Facts abduced from the knowledge base of $\HED$ with number base $2$.     }
\label{fig:kb-hed-2-facts}
\end{figure*}

\begin{figure*}[!h]
    \centering
    \fbox{
    \begin{minipage}{0.4\linewidth}
    \small
        \texttt{zero([Y0,Y1]) $\gets$ 0 is Y0 $+$ Y1. \\
        one([Y0,Y1]) $\gets$ 1 is Y0 $+$ Y1. \\
        two([Y0,Y1]) $\gets$ 2 is Y0 $+$ Y1. \\
        $\cdots \cdots$ \\
        seventeen([Y0,Y1]) $\gets$ 17 is Y0 $+$ Y1. \\
        eighteen([Y0,Y1]) $\gets$ 18 is Y0 $+$ Y1. 
        }
    \end{minipage}
    } 
\caption{A knowledge base of $\Addition$.    }
\label{fig:kb-addition}
\end{figure*}

\begin{figure*}[!h]
    \centering
    \fbox{
    \begin{minipage}{0.4\linewidth}
    \small
        \texttt{zero([0,0]). \\
        one([0,1]). \\ one([1,0]).\\
        two([0,2]). \\ two([1,1]). \\ two([2,0]).\\
        $\cdots \cdots$ \\
        seventeen([8,9]). \\ seventeen([9,8]).\\
        eighteen([9,9]). 
        }
    \end{minipage}
    }
\caption{Facts abduced from the knowledge base of $\Addition$ in \cref{fig:kb-addition}.     }
\label{fig:kb-addition-facts}
\end{figure*}

\clearpage

\section{Further Results on Benchmark Tasks}
\label{app:further-results}

The observations reported in the main text are further corroborated by \cref{tab:exp-resnet}, which showcases the test performance of ResNet-18 \cite{he2016deep} produced by different methods on different datasets and tasks.

\begin{table*}[!h]
\centering
\begin{small}
\resizebox{1.0\textwidth}{!}{
\setlength{\tabcolsep}{3.6mm}{
\begin{sc}
\begin{tabular}{c|c|c|c|c|c|c}
\toprule
Task                            & Method     & MNIST             & EMNIST           & USPS             & Kuzushiji        & Fashion          \\ \midrule
\multirow{5}{*}{$\ConjEq$}      & $\ABLRAND$ & $99.82 \pm 0.09$  & $99.54 \pm 0.10$ & $98.85 \pm 0.37$ & $98.57 \pm 0.14$ & $98.57 \pm 0.28$ \\
                                & $\ABLMAXP$ & $99.99 \pm 0.02$  & $99.83 \pm 0.08$ & $99.30 \pm 0.32$ & $99.65 \pm 0.12$ & $99.63 \pm 0.08$ \\
                                & $\ABLMIND$ & $99.89 \pm 0.06$  & $99.61 \pm 0.13$ & $99.36 \pm 0.27$ & $99.27 \pm 0.18$ & $99.06 \pm 0.30$ \\
                                & $\ABLAVG$  & $99.94 \pm 0.06$  & $99.69 \pm 0.18$ & $98.88 \pm 0.78$ & $98.33 \pm 0.38$ & $99.15 \pm 0.22$ \\
                                & $\ABLTL$   & $99.93 \pm 0.05$  & $99.83 \pm 0.08$ & $99.33 \pm 0.19$ & $99.58 \pm 0.07$ & $99.53 \pm 0.11$ \\ \midrule
\multirow{5}{*}{$\Conjunction$} & $\ABLRAND$ & $99.98 \pm 0.02$  & $99.98 \pm 0.04$ & $99.36 \pm 0.10$ & $99.38 \pm 0.33$ & $99.58 \pm 0.08$ \\
                                & $\ABLMAXP$ & $99.98 \pm 0.02$  & $99.91 \pm 0.08$ & $99.46 \pm 0.08$ & $99.62 \pm 0.10$ & $99.68 \pm 0.06$ \\
                                & $\ABLMIND$ & $100.00 \pm 0.00$ & $99.93 \pm 0.06$ & $99.44 \pm 0.09$ & $99.62 \pm 0.05$ & $99.71 \pm 0.05$ \\
                                & $\ABLAVG$  & $99.98 \pm 0.03$  & $99.95 \pm 0.03$ & $99.25 \pm 0.13$ & $99.57 \pm 0.11$ & $99.48 \pm 0.16$ \\
                                & $\ABLTL$   & $99.97 \pm 0.02$  & $99.95 \pm 0.04$ & $99.36 \pm 0.10$ & $99.70 \pm 0.09$ & $99.70 \pm 0.12$ \\ \midrule
\multirow{5}{*}{$\Addition$}    & $\ABLRAND$ & $98.76 \pm 0.31$  & $98.36 \pm 0.32$ & $95.59 \pm 1.06$ & $91.63 \pm 1.75$ & $88.44 \pm 0.78$ \\
                                & $\ABLMAXP$ & $99.57 \pm 0.03$  & $99.65 \pm 0.06$ & $97.28 \pm 0.14$ & $96.38 \pm 3.95$ & $91.08 \pm 4.26$ \\
                                & $\ABLMIND$ & $99.58 \pm 0.03$  & $99.67 \pm 0.03$ & $97.49 \pm 0.15$ & $98.15 \pm 0.08$ & $92.99 \pm 0.25$ \\
                                & $\ABLAVG$  & $99.10 \pm 0.09$  & $99.10 \pm 0.10$ & $96.45 \pm 0.38$ & $94.64 \pm 0.89$ & $90.35 \pm 0.24$ \\
                                & $\ABLTL$   & $99.39 \pm 0.08$  & $99.57 \pm 0.07$ & $97.37 \pm 0.07$ & $97.49 \pm 0.28$ & $92.38 \pm 0.18$ \\ \midrule
\multirow{5}{*}{$\HED$}         & $\ABLRAND$ & $99.93 \pm 0.04$  & $99.82 \pm 0.08$ & $99.41 \pm 0.31$ & $98.96 \pm 0.26$ & $99.12 \pm 0.34$ \\
                                & $\ABLMAXP$ & $100.00 \pm 0.00$ & $99.91 \pm 0.04$ & $99.66 \pm 0.00$ & $99.64 \pm 0.07$ & $99.57 \pm 0.09$ \\
                                & $\ABLMIND$ & $100.00 \pm 0.00$ & $99.90 \pm 0.06$ & $99.66 \pm 0.00$ & $97.81 \pm 3.94$ & $99.63 \pm 0.02$ \\
                                & $\ABLAVG$  & $99.53 \pm 0.14$  & $99.03 \pm 0.65$ & $99.57 \pm 0.15$ & $99.01 \pm 0.16$ & $98.32 \pm 0.52$ \\
                                & $\ABLTL$   & $100.00 \pm 0.00$ & $99.96 \pm 0.02$ & $99.66 \pm 0.00$ & $99.56 \pm 0.25$ & $99.59 \pm 0.17$ \\ \bottomrule
\end{tabular}
\end{sc}
}
}
\end{small}
\caption{Test accuracy (\%) of each method using ResNet-18 \cite{he2016deep} on benchmark datasets and tasks.}
\label{tab:exp-resnet}
\end{table*}

\section{Examples of Random Knowledge Bases}
\label{app:random-knowledge-bases}

\begin{figure*}[!h]
    \centering
    \fbox{
    \begin{minipage}{0.83\linewidth}
    \small
        \texttt{positive([Y0,Y1,Y2]) $\gets$ (Y0$\wedge$$\neg$Y1$\wedge$Y2)$\vee$($\neg$Y0$\wedge$$\neg$Y1$\wedge$Y2)$\vee$($\neg$Y0$\wedge$Y1$\wedge$$\neg$Y2). \\
        negative([Y0,Y1,Y2]) $\gets$ $\neg$ positive([Y0,Y1,Y2]).
        }
    \end{minipage}
    }
\caption{A randomly generated knowledge base in $\DNF$ that satisfies the rank criterion.   }
\end{figure*}

\begin{figure*}[!h]
    \centering
    \fbox{
    \begin{minipage}{0.83\linewidth}
    \small
        \texttt{positive([Y0,Y1,Y2]) $\gets$ ($\neg$Y0$\wedge$$\neg$Y1$\wedge$Y2)$\vee$(Y0$\wedge$$\neg$Y1$\wedge$$\neg$Y2)$\vee$(Y0$\wedge$Y1$\wedge$Y2)$\vee$($\neg$Y0$\wedge$Y1$\wedge$$\neg$Y2). \\
        negative([Y0,Y1,Y2]) $\gets$ $\neg$ positive([Y0,Y1,Y2]).
        }
    \end{minipage}
    }
\caption{A randomly generated knowledge base in $\DNF$ that does not satisfy the rank criterion.   }
\end{figure*}

\begin{figure*}[!h]
    \centering
    \fbox{
    \begin{minipage}{0.83\linewidth}
    \small
        \texttt{positive([Y0,Y1,Y2]) $\gets$ ($\neg$Y0$\vee$$\neg$Y1$\vee$$\neg$Y2)$\wedge$($\neg$Y0$\vee$Y1$\vee$$\neg$Y2). \\
        negative([Y0,Y1,Y2]) $\gets$ $\neg$ positive([Y0,Y1,Y2]).
        }
    \end{minipage}
    }
\caption{A randomly generated knowledge base in $\CNF$ that satisfies the rank criterion.   }
\end{figure*}

\begin{figure*}[!h]
    \centering
    \fbox{
    \begin{minipage}{0.83\linewidth}
    \small
        \texttt{positive([Y0,Y1,Y2]) $\gets$ (Y0$\vee$$\neg$Y1$\vee$Y2)$\wedge$($\neg$Y0$\vee$Y1$\vee$$\neg$Y2). \\
        negative([Y0,Y1,Y2]) $\gets$ $\neg$ positive([Y0,Y1,Y2]).
        }
    \end{minipage}
    }
\caption{A randomly generated knowledge base in $\CNF$ that does not satisfy the rank criterion.   }
\end{figure*}

\end{document}